\documentclass{article} 
\usepackage{iclr2025_conference,times}
\iclrfinalcopy

\usepackage{amsmath,amsfonts,bm}

\def\eqref#1{equation~\ref{#1}}

\def\1{\bm{1}}

\DeclareMathAlphabet{\mathsfit}{\encodingdefault}{\sfdefault}{m}{sl}
\SetMathAlphabet{\mathsfit}{bold}{\encodingdefault}{\sfdefault}{bx}{n}

\usepackage[utf8]{inputenc} 
\usepackage[T1]{fontenc}    
\usepackage{url}            
\usepackage{booktabs}       
\usepackage{amsfonts}       
\usepackage{nicefrac}       
\usepackage{microtype}      
\usepackage{xcolor}         

\usepackage{graphicx}
\usepackage{subcaption}
\usepackage{makecell}
\usepackage{amsmath}
\usepackage{setspace}
\usepackage{colortbl}
\usepackage{longtable}

\usepackage{pifont}
\usepackage{multirow}
\usepackage{makecell}

\usepackage{changepage}
\usepackage{lipsum}
\usepackage[font={small}]{caption}
\usepackage{wrapfig}
\usepackage{titletoc}

\definecolor{fun}{RGB}{212,226,237}
\definecolor{geom}{RGB}{219,221,206}

\definecolor{citecolor}{RGB}{65,105,225}
\usepackage[pagebackref=true,breaklinks=true,letterpaper=true,colorlinks,
citecolor=citecolor,bookmarks=false]{hyperref}

\definecolor{colorbest}{RGB}{255,179,179}
\definecolor{colorsecond}{RGB}{165,192,230}
\definecolor{colorworstS}{RGB}{215,203,208}
\definecolor{colorworstX}{RGB}{174,201,197}
\newcommand{\best}[0]{\cellcolor{colorbest} }
\newcommand{\second}[0]{\cellcolor{colorsecond}}
\newcommand{\worstS}[0]{\cellcolor{colorworstS}}
\newcommand{\worstX}[0]{\cellcolor{colorworstX}}

\DeclareRobustCommand{\legendsquare}[1]{%
  \textcolor{#1}{\rule{2ex}{2ex}}%
}

\newcommand{\cmark}{\ding{51}}
\newcommand{\xmark}{{\color{lightgray}\ding{55}}}

\title{\raisebox{-0.15\height}{\includegraphics[width=0.25in]{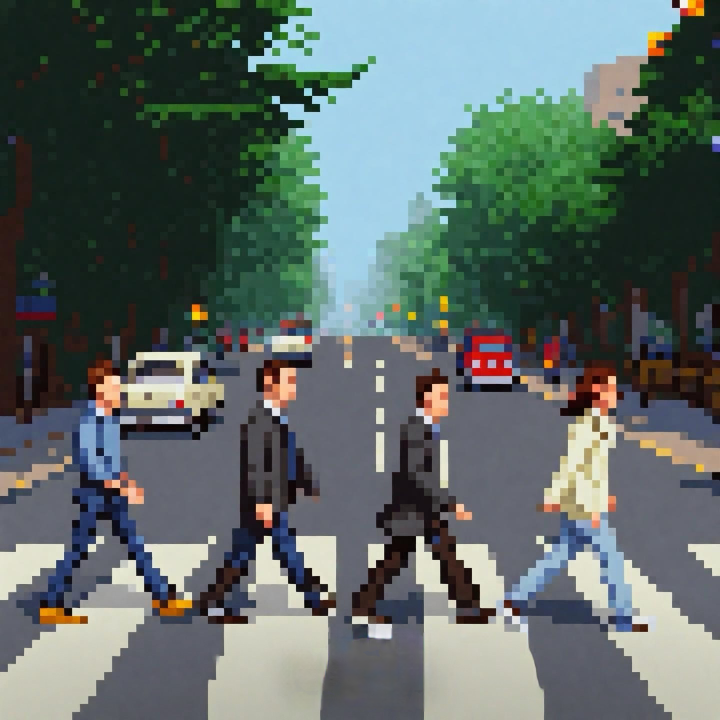}}
MetaUrban: An Embodied AI Simulation\\Platform for Urban Micromobility}

\author{Wayne Wu$^*$, Honglin He$^*$, Jack He, Yiran Wang, \\
    \textbf{Chenda Duan, Zhizheng Liu, Quanyi Li, Bolei Zhou} \\
    University of California, Los Angeles \\
    \url{https://metadriverse.github.io/metaurban} \\
}

\begin{document}

\maketitle

\begin{figure*}[h!]
    \centering
    \vspace{-0.35in}
    \includegraphics[width=1\linewidth]{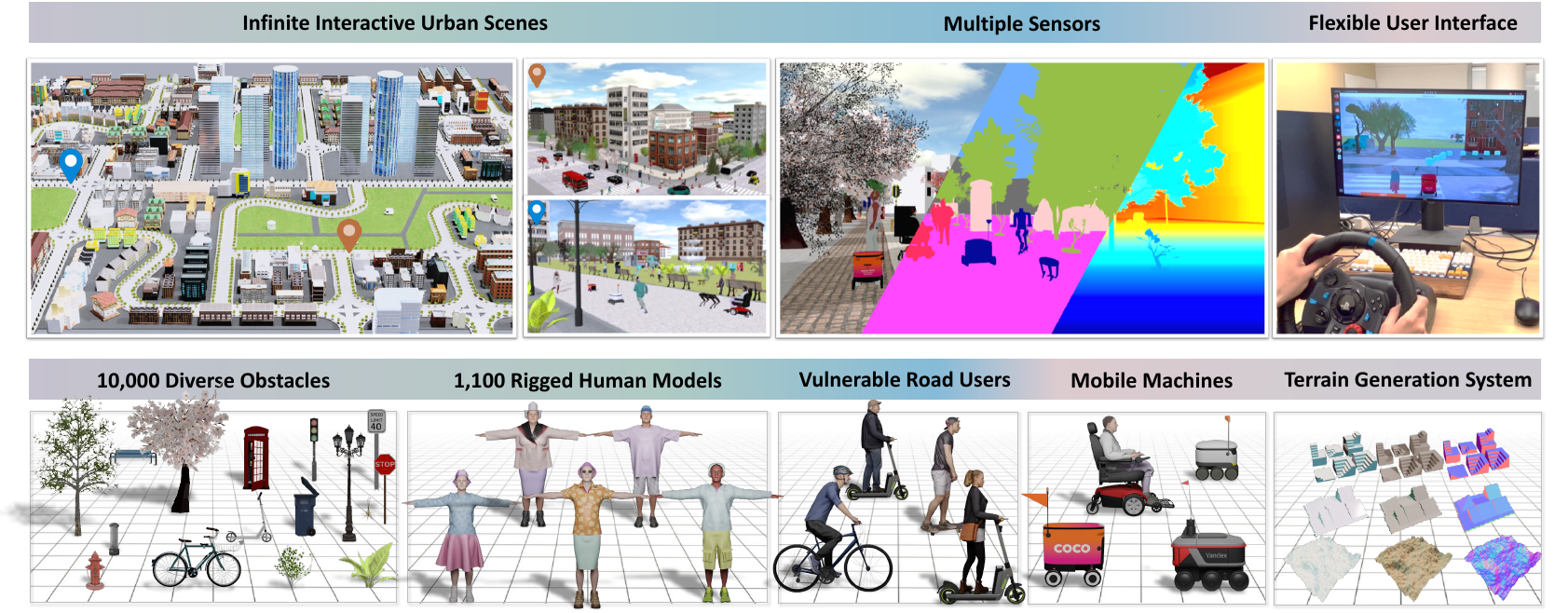}
    \caption{\textbf{MetaUrban} enables the construction of \textit{infinite interactive urban scenes}, supports \textit{multiple sensors}, and offers \textit{flexible user interfaces} such as a mouse, keyboard, joystick, and racing wheel. The platform includes \textit{10,000 diverse obstacles} in urban scenes, \textit{1,100 rigged human models} each with 2,314 movements, \textit{vulnerable road users}, \textit{mobile machines} with varied mechanical structures, and a \textit{terrain generation system} to create complex ground conditions. We highly recommend visiting our project page for video demonstrations.
    }
    \label{fig:teaser}
\end{figure*}

\begin{abstract}
\makeatletter
\def\blfootnote{\xdef\@thefnmark{}\@footnotetext}
\makeatother
\blfootnote{$*$ Equal contribution.}

Public urban spaces like streetscapes and plazas serve residents and accommodate social life in all its vibrant variations.
Recent advances in Robotics and Embodied AI make public urban spaces no longer exclusive to humans. Food delivery bots and electric wheelchairs have started sharing sidewalks with pedestrians, while robot dogs and humanoids have recently emerged in the street.
\textbf{Micromobility} enabled by AI for short-distance travel in public urban spaces plays a crucial component in the future transportation system.
Ensuring the generalizability and safety of AI models maneuvering mobile machines is essential.
In this work, we present \textbf{MetaUrban}, a \textit{compositional} simulation platform for the AI-driven urban micromobility research.
MetaUrban can construct an \textit{infinite} number of interactive urban scenes from compositional elements, covering a vast array of ground plans, object placements, pedestrians, vulnerable road users, and other mobile agents' appearances and dynamics.
We design point navigation and social navigation tasks as the pilot study using MetaUrban for urban micromobility research and establish various baselines of Reinforcement Learning and Imitation Learning.
We conduct extensive evaluation across mobile machines, demonstrating that heterogeneous mechanical structures significantly influence the learning and execution of AI policies.
We perform a thorough ablation study, showing that the compositional nature of the simulated environments can substantially improve the generalizability and safety of the trained mobile agents.
MetaUrban will be made publicly available to provide research opportunities and foster safe and trustworthy embodied AI and micromobility in cities. The code and dataset will be released.

\end{abstract}

\vspace{-0.15in}

\section{Introduction}
\label{sec:introduction}
\vspace{-0.15in}

Public urban spaces~\citep{whyte2012city} vary widely in type, form, and size, encompassing streetscapes, plazas, and parks. They are crucial spaces for transit and transport~\citep{geddes1949cities}, as well as providing stages to host various social events~\citep{park1925city}. In recent years, these spaces have also become key zones for the growing trend of \textbf{micromobility}~\citep{mitchell2010reinventing,abduljabbar2021role,oeschger2020micromobility}, a term that refers to small, lightweight vehicles like electric scooters, e-bikes, and other mobile machines designed for short-distance travel. Micromobility is becoming an increasingly important solution for improving urban transportation efficiency, reducing environmental impact, and offering flexible alternatives to car ownership in cities.

\begin{figure*}[h!]
    \centering
    \vspace{-0.12in}
    \includegraphics[width=1\linewidth]{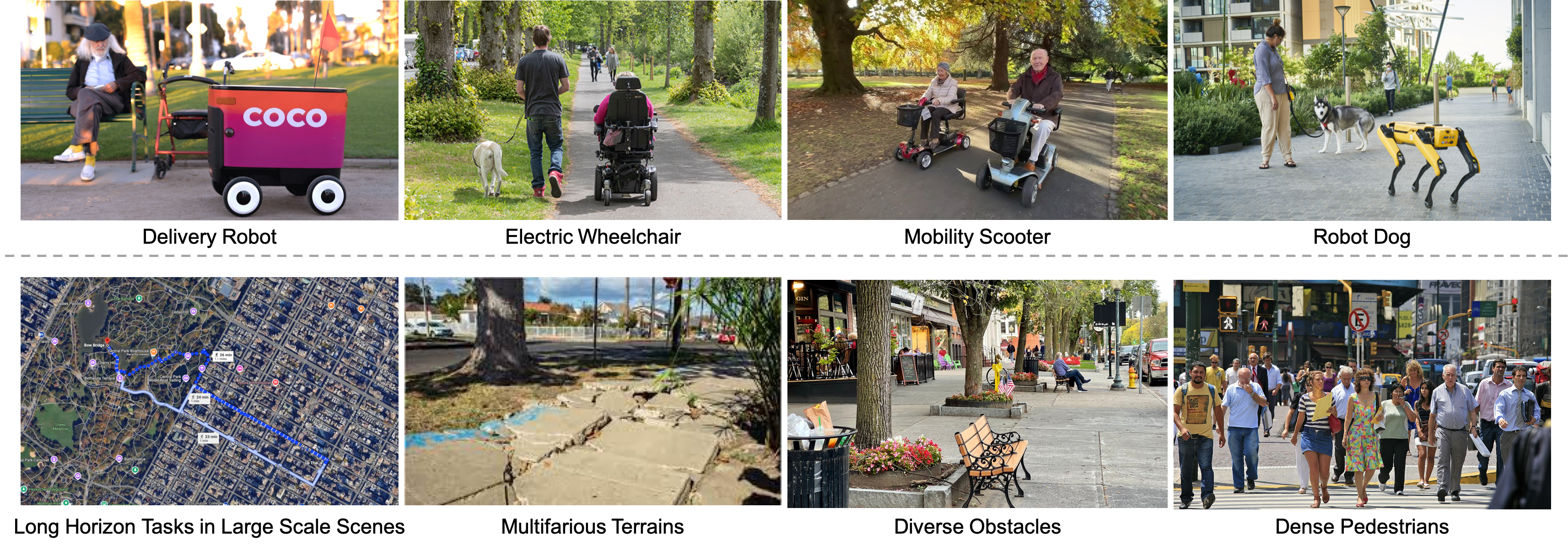}
    \vspace{-0.15in}
    \caption{\textbf{Motivation.} (Top) Emerging automated micromobility. (Bottom) Unique challenges in micromobility.
    }
    \vspace{-0.1in}
    \label{fig:intro}
\end{figure*}

As shown in Figure~\ref{fig:intro} (Top), food delivery bots navigate on the sidewalk to accomplish the last-mile food delivery task, while elders and physically disabled people maneuver electronic wheelchairs and mobility scooters on the street.
Various legged robots like robot dog Spot from Boston Dynamics and humanoid robot Optimus from Tesla are also forthcoming.
We can imagine a future of such \textit{automated micromobility} that harnesses advanced AI models to improve situational awareness and maneuver various mobile machines more intelligently and safely in complex urban environments.

Simulation platforms have played a crucial role in enabling systematic and scalable training of embodied AI agents and safety evaluation before real-world deployment. However, most of the existing simulators focus either on \textit{indoor household environments}~\citep{puig2018virtualhome,kolve2017ai2,savva2019habitat,shen2021igibson,li2024behavior,gan2020threedworld} or \textit{outdoor driving environments}~\citep{krajzewicz2002sumo,li2022metadrive,dosovitskiy2017carla}.
For example, platforms like AI2-THOR~\citep{kolve2017ai2}, Habitat~\citep{savva2019habitat}, and iGibson~\citep{shen2021igibson} are designed for household assistant robots in which the environments are mainly apartments or houses with furniture and appliances; platforms like SUMO~\citep{krajzewicz2002sumo}, CARLA~\citep{dosovitskiy2017carla}, and MetaDrive~\citep{li2022metadrive} are designed for research on autonomous driving and transportation focusing on roadways and highways.
Yet, simulating complex \textit{urban environments} for micromobility tasks, with diverse layouts, terrains, obstacles, and complex dynamics of pedestrians, is much less explored.

Distinct from the indoor household and driving tasks, micromobility plays an essential role in providing accessibility (\textit{e.g.}, electric wheelchairs) and convenience (\textit{e.g.}, food delivery bots) in the public urban space, while it also brings \textbf{\textit{unique challenges}} for mobile machines and the underlying embodied AI agents.
Let's follow the adventure of a last-mile delivery bot, who aims to deliver a lunch order from a nearby pizzeria to the campus (Figure~\ref{fig:intro} (Bottom)).
First, it faces a long-horizon task in \textbf{large scale} scenes across several street blocks, which span a significantly larger space than the indoor household environment.
Second, it needs to deal with \textbf{multifarious terrains}, such as fragmented curbs and rugged ground caused by tree roots on sidewalks, which are seldom seen in indoor and driving environments.
Then, it must safely navigate the cluttered street full of \textbf{diverse obstacles} like trash bins, parked scooters, and potted plants, which is absent in driving scenarios while with large obstacles variations compared to indoor scenes.
In addition, it needs to handle \textbf{dense pedestrians} on sidewalks to avoid collisions, especially taking care of disabled people in wheelchairs, which do not exist in indoor and driving environments.
Thus, the large scales, multifarious terrains, diverse obstacles, and dense pedestrians bring unique challenges to AI-driven mobile machines moving in cities, as well as the design of simulation environments for the training and evaluation of the embodied AI models.
In this work, we present \textbf{MetaUrban} -- a compositional simulation platform aiming to facilitate the research of AI-driven micromobility.
First, we introduce \textit{Hierarchical Layout Generation}, a procedural generation approach that can generate infinite layouts hierarchically from street blocks to sidewalks, functional zones, and object locations. It can generate scenes at an arbitrary scale with various connections and divisions of street blocks, obstacle locations, and complex terrains.
Then, we design the \textit{Scalable Obstacle Retrieval}, an automatic pipeline for acquiring an arbitrary number of high-quality objects with real-world distribution, to fill the urban space.
We first compute the object category distribution from worldwide urban scene data to form a description pool. Then, with the sampled descriptions from the pool, we design a VLM-based open-vocabulary searching schema, which can effectively retrieve objects from large-scale 3D asset repositories.
These two modules are critical for improving the \textit{generalizability} of trained agents.

Finally, we propose the \textit{Cohabitant Populating} method to generate complex dynamics in urban spaces. We first tailor recent 3D digital human and motion datasets to get 1,100 rigged human models, each with 2,314 movements.
Then, to form safety-critical scenarios, we integrate Vulnerable Road Users (VRUs) like bikers, skateboarders, and scooter riders.
As the subjects of micromobility, we include various mobile machines -- the delivery bot, electric wheelchair, mobility scooter, robot dog, and humanoid robot. Then, based on path planning algorithms, we can get complex trajectories among hundreds of environmental agents simultaneously with collision and deadlock avoidance.
Also, enabled by MetaUrban's flexible user interfaces (mouse, keyboard, joystick, and racing wheel), users can directly apply human-operated trajectories to agents, which provides an easy way to collect demonstration data for agent training.
In addition, we impose a series of traffic rules to regulate all agent behaviors.
It is critical for enhancing the \textit{safety} of the mobile agents.

Based on MetaUrban, we construct a large-scale dataset, MetaUrban-12K, that includes 12,800 training scenes and 1,000 test scenes. The mean area size is 20,000$m^2$, while the episode length is 410$m$ on average.
As a pilot study, we introduce Point Navigation and Social Navigation, which are the two most fundamental tasks for mobile machines moving in urban spaces, as a starting point for AI-driven micromobility research.
We build comprehensive benchmarks for these two tasks, in which we establish extensive baseline models, covering Reinforcement Learning, Safe Reinforcement Learning, Offline Reinforcement Learning, and Imitation Learning.
Then, we make extensive evaluations across mobile machines to delve into the performance influence of varied mechanical structures (such as engine force, wheel friction, and wheelbase) on the learning and execution of AI policies.
In the ablation study, we demonstrate that the compositional nature of the simulated environments can substantially improve the generalizability and safety of the trained mobile agents.
We will make MetaUrban publicly available to enable more research opportunities for the community and foster safe and trustworthy embodied AI and micromobility in cities.

\vspace{-0.15in}
\section{Related Work}
\label{sec:related_work}
\vspace{-0.15in}

Many simulation platforms have been developed for embodied AI research, depending on the target environments -- such as indoor homes and offices, driving roadways and highways, and crowds in warehouses and squares.
We compare representative ones with the proposed MetaUrban simulator.

\vspace{-0.12in}
\paragraph{Indoor Environments.}
Platforms for indoor environments are mainly designed for household assistant robots, emphasizing the affordance, realism, and diversity of objects, as well as the interactivity of environments.
VirtualHome~\citep{puig2018virtualhome} pivots towards simulating routine human activities at home.
AI2-THOR~\citep{kolve2017ai2} and its extensions, such as ManipulaTHOR~\citep{ehsani2021manipulathor}, RoboTHOR~\citep{deitke2020robothor}, and ProcTHOR~\citep{deitke2022procthor}, focus on detailed agent-object interactions, dynamic object state changes, and procedural scene generation, alongside robust physics simulations.
Habitat~\citep{savva2019habitat} offers environments reconstructed from 3D scans of real-world interiors. Its subsequent iterations, Habitat 2.0~\citep{szot2021habitat} and Habitat 3.0~\citep{puig2023habitat}, introduce interactable objects and deformable humanoid agents, respectively.
iGibson~\citep{shen2021igibson} provides photorealistic environments. Its upgrades, Gibson 2.0~\citep{li2021igibson}, and  OmniGibson~\citep{li2024behavior}, focus on household tasks with object state changes and a realistic physics simulation of everyday activities, respectively.
ThreeDWorld~\citep{gan2020threedworld} targets real-world physics by integrating high-fidelity simulations of liquids and deformable objects.
However, unlike MetaUrban, these simulators are focused on indoor environments with particular tasks like object rearrangement and manipulation.

\vspace{-0.12in}
\paragraph{Driving Environments.}
Platforms for driving environments are mainly designed for autonomous vehicle research and development.
Simulators like GTA V~\citep{martinez2017beyond}, Sim4CV~\citep{muller2018sim4cv}, AIRSIM~\citep{shah2018airsim}, CARLA~\citep{dosovitskiy2017carla}, and its extension SUMMIT~\citep{cai2020summit} offer realistic environments that mimic the physical world’s detailed visuals, weather conditions, and day-to-night transitions.
Other simulators enhance efficiency and extensibility at the expense of visual realism, such as Udacity~\citep{udacity}, DeepDrive~\citep{deepdrive}, Highway-env~\citep{highway-env}, and DriverGym~\citep{kothari2021drivergym}.
MetaDrive~\citep{li2022metadrive} trades off between visual quality and efficiency, offering a lightweight driving simulator that can support the research of generalizable RL algorithms for vehicles.
Although some of the simulators~\citep{martinez2017beyond,dosovitskiy2017carla} involve traffic participants other than vehicles, such as pedestrians and cyclists, all of them focus on vehicle-centric driving scenarios and neglect the stage for urban micromobility -- public urban spaces like sidewalks and plazas.

\vspace{-0.12in}
\paragraph{Social Navigation Environments.}
Other than indoor and driving environments, social navigation platforms emphasize the social compatibility of robots.
Simulators like Crowd-Nav~\citep{chen2019crowd}, Gym-Collision-Avoidance~\citep{everett18motion}, and Social-Gym 2.0~\citep{sprague2023socialgym}, model scenes and agents in 2D maps, focusing more on the development of path planning algorithms.
Other simulators, such as HuNavSim~\citep{perez2023hunavsim}, SEAN 2.0~\citep{tsoi2022sean}, and SocNavBench~\citep{biswas2022socnavbench}, upgrade the environment to 3D space and introduce human pedestrians to support the development of more complex algorithms.
Social navigation platforms focus on crowd navigation, with oversimplified objects and surrounding environmental structures in the scenes, making them not applicable to complex urban micromobility tasks.
In contrast, MetaUrban supports large-scale urban space simulation with real-world scenes (such as street facilities and terrains), providing significantly rich semantics and superior complexity of environments. In addition, MetaUrban supports the cross-machine evaluation of generalizability and safety with different mechanical structures. These features make it a unique choice for urban micromobility.

In summary, none of the recent simulation platforms have been constructed for urban micromobility. The proposed simulator MetaUrban is the first simulator designed for AI-driven urban micromobility research. It differs from previous simulators significantly in terms of complex scenes (with large scales and multifarious terrains), diverse obstacles, vibrant dynamics, different types of mobile machines like delivery bots, electric wheelchairs, and mobility scooters, and intricate interactions simulated.
Please refer to Appendix~\ref{sec:simulator_comparison} for a detailed comparison table with existing simulators in the dimensions of scale, sensor, and features, where MetaUrban shows a remarkable superiority.
We believe MetaUrban can provide a lot of new research opportunities for AI-driven urban micromobility.

\vspace{-0.15in}
\section{MetaUrban Simulator}
\vspace{-0.1in}

\begin{figure*}[h!]
    \centering
    \vspace{-0.1in}
    \includegraphics[width=1\linewidth]{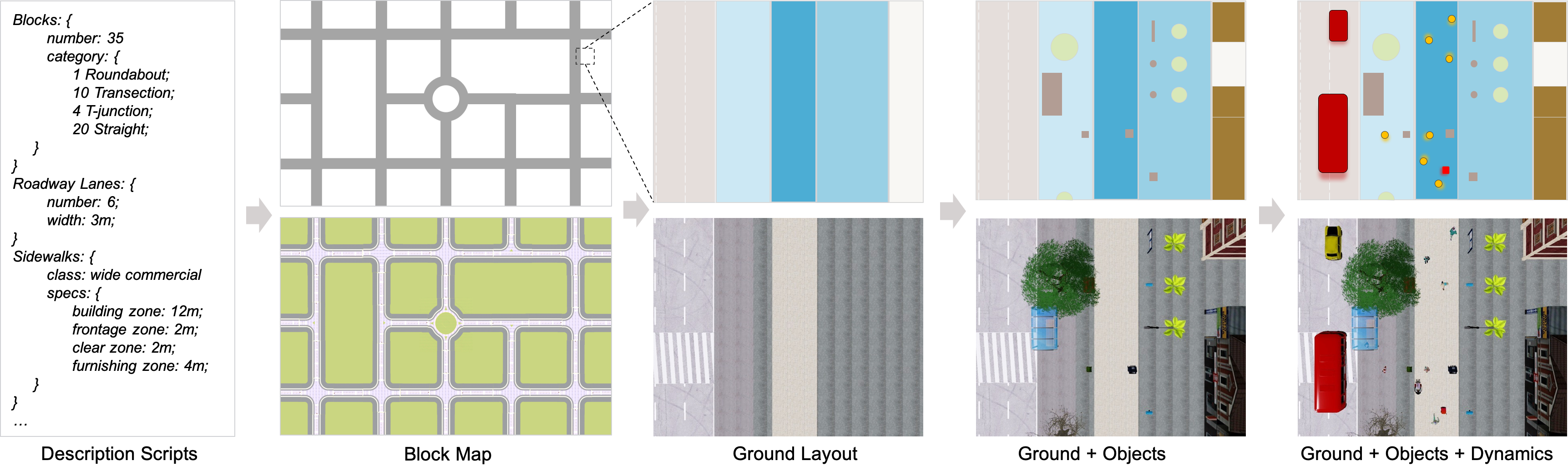}
    \vspace{-0.15in}
    \caption{\textbf{Procedural generation.} MetaUrban can automatically generate complex urban scenes with its compositional nature. From the second to the fourth column, the top row shows the 2D road maps, and the bottom row shows the bird-eye view of 3D scenes.
    }
    \vspace{-0.12in}
    \label{fig:procedural_generation}
\end{figure*}

MetaUrban is a compositional simulation platform that can generate infinite training and evaluation environments for AI-driven urban micromobility.
We propose a procedural generation pipeline, as the basis of MetaUrban, for constructing infinite interactive scenes with different specifications. As shown in Figure~\ref{fig:procedural_generation}, MetaUrban uses a structured description script to create urban scenes. Based on the script information about street blocks, sidewalks, objects, agents, and more, it starts with the street block map, then plans the ground layout by dividing different function zones, then places static objects, and finally populates dynamic agents.

This section highlights three key designs in the MetaUrban simulator to support exhibiting three unique characteristics of urban spaces respectively -- complex scenes (with large scales and multifarious terrains), diverse obstacles, and vibrant dynamics.
Section~\ref{sec:layout_generation} introduces \textbf{Hierarchical Layout Generation}, which can infinitely generate diverse layouts with different functional zone divisions, object locations, and terrains that are essential for the \textit{generalizability} to scene diversity of agents.
Section~\ref{sec:assets_retrival} introduces \textbf{Scalable Obstacle Retrieval}, which harnesses worldwide urban scene data to obtain real-world object distributions in different places, and then builds large-scale, high-quality static objects set with VLM-enabled open-vocabulary searching. It is crucial for further enhancing the \textit{generalizability} to obstacle diversity of agents.
Section~\ref{sec:co_inhabitant_populating} introduces \textbf{Cohabitant Populating}, in which we leverage the advancements in digital humans to enrich the appearances, movements, and trajectories of pedestrians and vulnerable road users, as well as incorporate other agents to form a vivid cohabiting environment. It is critical for improving the \textit{safety} of the mobile agents.

\vspace{-0.13in}
\subsection{Hierarchical Layout Generation}
\label{sec:layout_generation}
\vspace{-0.1in}

The complexity of scene layouts, \textit{i.e.}, the connection and categories of blocks, the specifications of sidewalks and crosswalks, the placement of objects, as well as the terrains, is crucial for enhancing the generalizability of trained agents maneuvering in public spaces. In the hierarchical layout generation framework, we start with a ground plan that samples categories of street blocks and divides sidewalks into different functional zones. Then, we allocate various objects procedurally conditioned on functional zones. Finally, we implement a terrain generation system to synthesize various ground conditions. With the above procedures, we can get infinite urban scene layouts with different specifications of sizes, object locations, and terrains.

\vspace{-0.02in}
\noindent
\textbf{Ground plan.}
We design 5 typical street block categories, \textit{i.e.}, straight, intersection, roundabout, circle, and T-junction. In the simulator, to form a large map with several blocks, we can sample the category, number, and order of blocks, as well as the number and width of lanes in one block, to get different maps.
Then, each block can simulate its own walkable areas -- sidewalks and crosswalks, which are key areas for urban spaces with plenty of interactions.

\begin{wrapfigure}{r}{0.6\textwidth}
  \begin{center}
\vspace{-0.25in}
\includegraphics[width=1\linewidth]{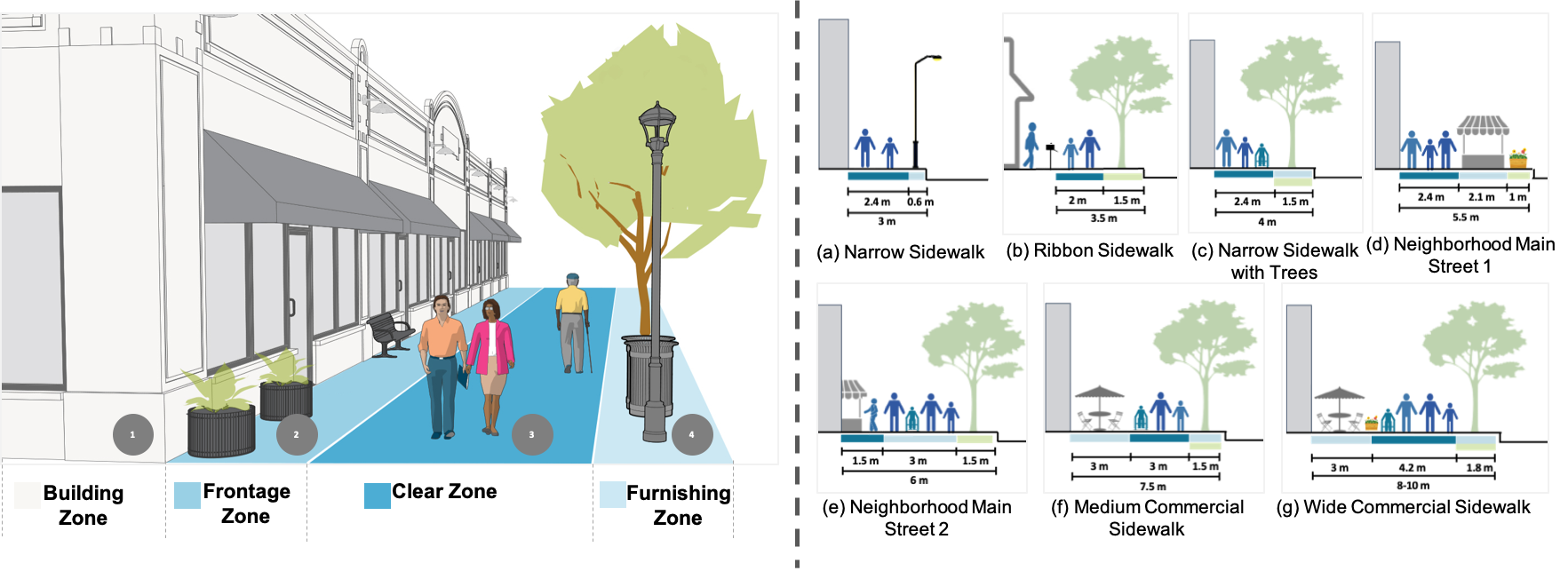}
\vspace{-5mm}
    \caption{\textbf{Ground plan.} (Left) Sidewalk is divided into four functional zones -- building, frontage, clear, and furnishing zone. (Right) Seven typical sidewalk templates -- from (a) to (g).}
    \vspace{-0.21in}
    \label{fig:ground_plan}
    \end{center}
\end{wrapfigure}

As shown in Figure~\ref{fig:ground_plan} (Left), according to the Global Street Design Guide~\citep{global2016global} provided by the Global Designing Cities Initiative,
we divide the sidewalk into four functional zones -- building zone, frontage zone, clear zone, and furnishing zone. Based on their different combinations of functional zones, we further construct 7 typical templates for sidewalks (Figure~\ref{fig:ground_plan} (Right)). To form a sidewalk, we can first sample the layout from the templates and then assign proportions for different function zones.
For crosswalks, we provide candidates at the start and the end of each roadway, which support specifying the needed crosswalks or sampling them by a density parameter.

\vspace{-0.02in}
\noindent
\textbf{Terrain generation.}
We develop a procedural terrain generator by connecting sampled terrain primitives similar to the method adopted in~\citep{lee2024learning}, which uses the Wave Function Collapse (WFC) method~\citep{gumin2016wave} to ensure smooth transitions between neighboring terrain primitives. We defined five types of terrain primitives, including slops, steps, stairs, ramps, and rough at different heights. After the mesh was generated, textures with different friction coefficients
were added to the terrain to simulate different materials of the ground. 
This method allows for the generation of a wide variety of terrain combinations, reflecting the complex environments that agents may encounter.

\vspace{-0.02in}
\noindent
\textbf{Object placement.}
After determining the ground layout, we can place objects on the ground. We divide objects into three classes. 1) Standard infrastructure, such as poles, trees, and signs, are placed periodically along the road. 2) Non-standard infrastructure, such as buildings, bonsai, and trash bins, are placed randomly in the designated function zones. 3) Clutter, such as drink cans, bags, and bicycles, are placed randomly across all functional zones.
We can get different street styles by specifying an object pool while getting different compactness by specifying a density parameter.
Please refer to Appendix~\ref{sec:layoutgeneration} for more details.

\vspace{-0.15in}
\subsection{Scalable Obstacle Retrieval}
\label{sec:assets_retrival}
\vspace{-0.1in}

Hierarchical layout generation decides the scene's layout and \textit{where} to place the objects. However, to make the trained agents generalizable when navigating through scenes composed of various objects in the real world, \textit{what} objects to place is another crucial question.
In this section, we propose the Scalable Obstacle Retrieval pipeline, in which we first get real-world object distributions from web data, and then retrieve objects from 3D asset repositories through an open-vocabulary search schema based on VLMs.
This pipeline is flexible and extensible: the retrieved objects can be scaled to arbitrary sizes as we continue to exploit more web data for scene descriptions and include more 3D assets as the candidate objects.

\begin{figure*}[t]
    \centering
    \includegraphics[width=1\linewidth]{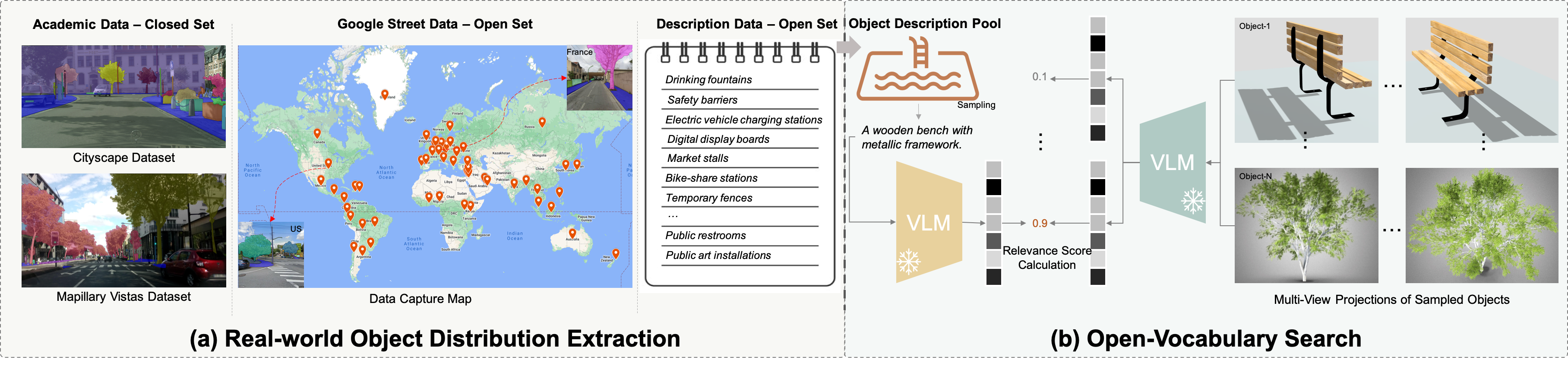}
    \vspace{-0.2in}
    \caption{\textbf{Scalable obstacle retrieval.} (a) Real-world distribution extraction. We get object distribution for urban spaces from three sources: academic datasets, Google Street data, and text description data. (b) Open-vocabulary search. We use the VLM to get image and text embedding, respectively. Then, based on the relevant scores, we can get the objects with high rankings.
    }
    \vspace{-0.2in}
    \label{fig:assets_retrieval}
\end{figure*}

\vspace{-0.02in}
\noindent
\textbf{Real-world object distribution extraction.}
Urban spaces have unique structures and object distributions, such as the infrastructure built by the urban planning administration and clutters placed by people. Thus, we design a real-world distribution extraction method to get a description pool depicting the frequent objects in urban spaces.
As illustrated in Figure~\ref{fig:assets_retrieval} (a), we first leverage off-the-shelf academic datasets for scene understanding, CityScape~\citep{cordts2016cityscapes} and Mapillary Vistas~\citep{neuhold2017mapillary}, to get a list of 90 objects that are with high frequency to be put in the urban space. However, the number of objects is limited because of the closed-set definitions in the image datasets. We introduce two open-set sources to get broader object distribution from the real world.
1) Google Street data. We first collect 25,000 urban space images from 50 countries across six continents. Then, we harness GPT-4o~\citep{gpt-4o} and open-set segmentation model Grounded-SAM~\citep{ren2024grounded} to get 1,075 descriptions of objects in the public urban space.
2) Urban planning description data. We further get a list of 50 essential objects in public urban spaces through a thorough survey of 10 urban design handbooks.
Finally, by combining these three data sources, we can get an object description pool with 1,215 items of descriptions that form the real-world object category distribution.

\vspace{-0.02in}
\noindent
\textbf{Open-vocabulary search.}
The recent development of large-scale 3D object repositories~\citep{deitke2023objaverse,deitke2024objaverse,wu2023omniobject3d} enables efficiently constructing a dataset for a specific scene. However, these large repositories have three intrinsic issues to harness these repositories: 1) most of the data is unrelated to the urban scene, 2) the data quality in large repositories is uneven, and 3) the data has no reliable attribute annotations.
To this end, we introduce an open-vocabulary search method to tackle these issues. As shown in Figure~\ref{fig:assets_retrieval} (b), the whole pipeline is based on an image-text retrieval architecture. We first sample objects from Objaverse~\citep{deitke2023objaverse} and Objaverse-XL~\citep{deitke2024objaverse} repositories to get projected multi-view images. Here, a naive uniform view sampling will bring low-quality harmful images. Following~\citep{luo2023scalable,luo2024view}, we select and prioritize informative viewpoints, which significantly enhance retrieval effectiveness. Then, we leverage the encoder of a Vision Language Model BLIP~\citep{li2022blip} to extract features from projected images and sampled descriptions from the object description pool, respectively, to calculate relevant scores. Then, we can get target objects with relevant scores up to a threshold.
This method lets us get an urban-specific dataset with 10,000 high-quality obstacles in real-world category distributions.

\vspace{-0.15in}
\subsection{Cohabitant Populating}
\label{sec:co_inhabitant_populating}
\vspace{-0.1in}

In this section, we will describe how to populate these static urban scenes with varied agents regarding appearances, movements, and trajectories through Cohabitant Populating.

Following BEDLAM~\cite{black2023bedlam} and AGORA~\cite{patel2021agora}, we represent humans as parametric human model SMPL-X~\cite{SMPL-X:2019}, in which the 3D human body is controlled by a set of parameters for pose $\theta$, shape $\beta$, and facial expression $\phi$, respectively. Then, built upon SynBody~\cite{yang2023synbody}'s asset repository, 1,100 3D rigged human models are procedually generated by sampling from 68 garments, 32 hairs, 13 beards, 46 accessories, and 1,038 cloth and skin textures.
To form safety-critical scenarios, we also include vulnerable road users like bikers, skateboarders, and scooter riders.
For the other agents, we incorporate the 3D assets of COCO Robotics', Starship’s and Yandex’s delivery bots, Drive Medical’s electric wheelchair, Pride Go-Go’s mobility scooter, Boston Dynamic’s robot dog, and Agility Robotics’ humanoid robot.

We provide two kinds of human movements in the simulator -- daily movements and unique movements. Daily movements provide the basic human dynamics in daily life, \textit{i.e.}, idle, walking, and running. Unique movements are the complicated dynamics that appear randomly in public spaces, such as dancing and exercising. We harness the BEDLAM dataset~\cite{black2023bedlam} to obtain 2,311 unique movements.
For humans and other agents with daily movements, we simulate their trajectories using the ORCA~\cite{van2011reciprocal} social forces model and Push and Rotate (P\&R) algorithm~\cite{de2014push}. ORCA~\cite{van2011reciprocal} uses a joint optimization and a centralized controller that guarantees that agents will not collide with each other or any other objects identified as obstacles. Push and Rotate (P\&R)~\cite{de2014push} is a multi-agent path-finding algorithm that can resolve any potential deadlock by local coordination.
In the future, an interesting direction is to endow personal traits like job, personality, and purpose to humans and harness the advantages of LLMs~\cite{achiam2023gpt} and LVMs~\cite{liu2023visual} to enable social~\cite{puig2023nopa} and interactive behaviors~\cite{park2023generative} of humans in urban scenes.
\vspace{-0.15in}
\section{Experiments}
\label{sec:experiments}
\vspace{-0.1in}

In this section, we first introduce the data and evaluation metrics used in all the experiments.
In Section~\ref{sec:benchmark}, we build comprehensive benchmarks for two core tasks in micromobility -- point navigation and social navigation, across seven typical baseline models.
In Section~\ref{sec:cross_machine_eva}, we make evaluations across different mobile machines to delve into the influence of mechanical structures, such as engine force, wheel friction, and wheelbase, on their capability when learning and executing policies.
In Section~\ref{sec:ablation_study}, we evaluate the generalizability and scaling ability of the MetaUrban platform, and reveal the effects of the density of static objects and dynamic agents.

\vspace{-0.12in}
\paragraph{Data.} 
Based on the MetaUrban simulator, we construct the MetaUrban-12K dataset, including 12,800 interactive urban scenes for training (\textit{MetaUrban-train}) and 1,000 scenes for testing (\textit{MetaUrban-test}).
Scenes in this dataset are connected by one to three street blocks covering an average of 20,000$m^2$ areas. There are an average of 0.03 static objects per $m^2$ and the average distance of objects is 0.7$m$. There are 10 dynamic agents per street block, including pedestrians, vulnerable road users, and other agents. The average episode length is 410$m$. These form significantly challenging scenes for agents to navigate through, which are crowded and have long horizons.
We further construct an unseen test set (\textit{MetaUrban-unseen}) with 100 scenes for the unseen evaluations.
To enable the fine-tuning experiments, we construct a training set of 1,000 scenes with the same distribution of MetaUrban-unseen, termed \textit{MetaUrban-finetune}.
Appendix~\ref{sec:metaurban_dataset} provides detailed descriptions, and statistics of the MetaUrban-12K dataset.

\vspace{-0.12in}
\paragraph{Evaluation metrics.}
The agent is evaluated using the Success Rate (SR) and Success weighted by Path Length (SPL)~\citep{anderson2018evaluation,batra2020objectnav} metrics, which measure the success and efficiency of the path taken by the agent. For SocialNav, except Success Rate (SR), the Social Navigation Score (SNS)~\citep{deitke2022retrospectives}, is also used to evaluate the social complicity of the agent. For both tasks, we further report the Cumulative Cost (CC)~\citep{li2022metadrive} to evaluate the safety properties of the agent. It records the crash frequency to obstacles or environmental agents.

\begin{table}[h]
\footnotesize
\vspace{-0.08in}
\caption{\textbf{Benchmarks.} The benchmark of PointNav and SocialNav tasks on the MetaUrban-12K dataset. Seven representative methods of RL, safe RL, offline RL, and IL are evaluated for each benchmark. \legendsquare{colorbest} indicate the best performance among online methods (RL and Safe RL). \legendsquare{colorsecond} indicates the best performance among offline methods (offline RL and IL).}
\vspace{-0.12in}
\label{tab:benchmark}
\begin{center}
\resizebox{1\columnwidth}{!}{
    \begin{tabular}{c|c|ccc|ccc|ccc|ccc}
        \toprule
        \multirow{3}{*}{\textbf{Category}} & \multirow{3}{*}{\textbf{Method}} & \multicolumn{6}{c|}{\textbf{PointNav}} & \multicolumn{6}{c}{\textbf{SocialNav}} \\
        \cmidrule{3-14}
        & & \multicolumn{3}{c|}{Test} & \multicolumn{3}{c|}{Unseen} & \multicolumn{3}{c|}{Test} & \multicolumn{3}{c}{Unseen} \\
        \cmidrule{3-14}
        & & \makecell{SR$\uparrow$} & \makecell{SPL$\uparrow$} & \makecell{Cost$\downarrow$} & \makecell{SR$\uparrow$} & \makecell{SPL$\uparrow$} & \makecell{Cost$\downarrow$} & \makecell{SR$\uparrow$} & \makecell{SNS$\uparrow$} & \makecell{Cost$\downarrow$} & \makecell{SR$\uparrow$} & \makecell{SNS$\uparrow$} & \makecell{Cost$\downarrow$} \\
        \toprule
        RL & PPO~\citep{schulman2017proximal} &\best{66\%} & \best{0.64} & 0.51 & 49\% & 0.45 & 0.78 & \best{34\%} & \best{0.64} & 0.66 & \best{24\%} & \best{0.57} & 0.51 \\
        \midrule
        \multirow{2}{*}{Safe RL} & PPO-Lag~\citep{ray2019benchmarking} & 60\% & 0.58 & \best{0.41} & \best{60\%} & \best{0.57} & \best{0.53} & 17\% & 0.51 & 0.33 & 8\% & 0.47 & \best{0.50} \\
        & PPO-ET~\citep{sun2021safe} & 57\% & 0.53 & 0.47 & 53\% & 0.49 & 0.65 & 5\% & 0.52 & \best{0.26} & 2\% & 0.50 & 0.62 \\
        \midrule
        \midrule
        \multirow{2}{*}{Offline RL} & IQL~\citep{kostrikov2021offline} & 36\% & 0.33 & \second{0.49} & 30\% & 0.27 & \second{0.63} & \second{36\%} & \second{0.67} & \second{0.39} & 27\% & 0.62 & 3.05 \\
        & TD3+BC~\citep{fujimoto2021minimalist} & 29\% & 0.28 & 0.77 & 20\% & 0.20 & 1.16 & 26\% & 0.61 & 0.62 & \second{32\%} & \second{0.64} & 1.53 \\
        \midrule
        \multirow{2}{*}{IL} & BC~\citep{michael1995bc} & 36\% & 0.28 & 0.83 & 32\% & 0.26 & 1.15 & 28\% & 0.56 & 1.23 & 18\% & 0.54 & \second{0.58} \\
        & GAIL~\citep{ho2016generative} & \second{47\%} & \second{0.36} & 1.05 & \second{40\%} & \second{0.32} & 1.46 & 34\% & 0.63 & 0.71 & 28\% & 0.61 & 0.67 \\
        \bottomrule
    \end{tabular}
}
\end{center}
\end{table}

\vspace{-0.2in}
\subsection{Benchmarks}
\label{sec:benchmark}
\vspace{-0.1in}

In micromobility, the core and foundational function for a mobile machine is \textit{to navigate from point A to point B within a bustling urban environment}~\citep{mitchell2010reinventing,gossling2020integrating}. In this function, there are two main commands: 1) not collide with static objects (infrastructures and clutters), and 2) not bump into moving objects (pedestrians and other agents). To this end, we design two foundational tasks -- point navigation and social navigation, to evaluate the generalizability and safety of recent state-of-the-art embodied AI models on a typical wheeled mobile machine\textsuperscript{\ref{fn:clarification}}, \textit{i.e.}, COCO Robotics Delivery Bot\textsuperscript{\ref{fn:coco_robot}}.

\footnotetext[3]{Note that MetaUrban provides an easy-to-use interface for users to benchmark the navigation ability of any mobile machine they want to evaluate, such as electric wheelchairs, robot dogs, and humanoids. In a navigation-locomotion framework~\citep{lee2024learning}, an off-the-shelf locomotion module can be fixed for one machine; users can change the navigation module independently to benchmark.\label{fn:clarification}}

\footnotetext[4]{\url{https://www.cocodelivery.com/}\label{fn:coco_robot}}

\vspace{-0.12in}
\paragraph{Tasks.}
We design two common tasks in urban scenes: Point Navigation (PointNav) and Social Navigation (SocialNav).
In PointNav, the agent’s goal is to navigate to the target coordinates in static environments without access to a pre-built environment map.
In SocialNav, the agent is required to reach a point goal in dynamic environments that contain moving environmental agents.
The agent shall avoid collisions or proximity to environmental agents beyond thresholds to avoid penalization (distance <0.2 meters).
The agent's action space in the experiments consists of acceleration, brake, and steering.
The observations contain a vector denoting the LiDAR signal, a vector summarizing the agent’s state, and the navigation information that guides the agent toward the destination.

\vspace{-0.12in}
\paragraph{Models.}
We build two benchmarks on the MetaUrban-12K dataset for PointNav and SocialNav tasks. We evaluate 7 typical baseline models, across Reinforcement Learning (PPO~\citep{schulman2017proximal}), Safe Reinforcement Learning (PPO-Lag~\citep{ray2019benchmarking}, and PPO-ET~\citep{sun2021safe}), Offline Reinforcement Learning (IQL~\citep{kostrikov2021offline} and TD3+BC~\citep{fujimoto2021minimalist}), and Imitation Learning (BC~\citep{michael1995bc} and GAIL~\citep{ho2016generative}).
We train all these baseline models on the MetaUrban-train dataset and then evaluate them on the MetaUrban-test set. We use the demonstration data provided in MetaUrban-12K for offline RL and IL training. We further make unseen evaluations on the MetaUrban-unseen set to demonstrate the generalizability of models trained on the MetaUrban-12K dataset while directly tested on unseen environments. 
Please refer to Appendix~\ref{sec:exp_detail} for details of models, rewards, and hyperparameters.

\vspace{-0.12in}
\paragraph{Results.}
Table~\ref{tab:benchmark} shows the results in the PointNav and SocialNav benchmarks.
We can draw two key observations from the results.
1) The tasks are far from being solved. The highest success rates are only $66\%$ and $36\%$ for PointNav and SocialNav tasks achieved by the baselines, indicating the difficulty of these tasks in the urban environments composed by MetaUrban. Note that these benchmarks are built on a medium level of object and dynamic density; increasing the density will further degrade the performances shown in ablation studies.
2) Models trained on MetaUrban-12K have strong generalizability in unseen environments. With unseen evaluation, models can still achieve $41\%$ and $26\%$ success rates on average for PointNav and SocialNav tasks. These results are strong since the models generalize to not only unseen objects and layouts but also unseen dynamics of environmental agents. It demonstrates that the compositional nature of MetaUrban, supporting the coverage of a large spectrum of complex urban scenes, can successfully empower generalization ability to the trained models.
In addition, other interesting observations include: SocialNav is much harder than PointNav due to the dynamics of the mobile environmental agents, and Safe RL remarkably improves the safety properties at the expense of effectiveness.

\vspace{-0.12in}
\subsection{Evaluation across Mobile Machines}
\label{sec:cross_machine_eva}
\vspace{-0.1in}

Different mobile machines have heterogeneous mechanical structure design specifications, which have a huge influence on their navigation behaviors and capabilities. MetaUrban can evaluate different mechanical structures and designs of mobile machines before deployment and volume production.
In this section, we evaluate three typical wheeled mobile machines (a delivery bot, an electric wheelchair, and a mobility scooter) with large design variations, and investigate the relationship between mechanical structures and performance in \textit{policy learning}. Further, we also investigate the influence of mechanical structures in \textit{policy execution} on different terrains, as detailed in Appendix~\ref{sec:policy_execution}.

\begin{table}[h!]
\small
\vspace{-0.12in}
\caption{\textbf{Evaluation of policy learning across mobile machines.} \legendsquare{colorbest} and \legendsquare{colorworstS} indicate the best and worst performance among machines in the straight street block (``S''). \legendsquare{colorsecond} and \legendsquare{colorworstX} indicate the best and worst performance among machines in the intersection street block (``X'').}
\vspace{-0.12in}
\label{tab:cross_machine_eva}
\begin{center}
\resizebox{1\columnwidth}{!}{
    \begin{tabular}{c|cc|cc|cc|cc|cc}
        \toprule
        &\multicolumn{2}{c|}{\textbf{COCO (base)}} & \multicolumn{2}{c|}{\textbf{COCO (mod-1)}} &  \multicolumn{2}{c|}{\textbf{COCO (mod-2)}} & \multicolumn{2}{c|}{\textbf{Wheelchair}} & \multicolumn{2}{c}{\textbf{Mobility Scooter}} \\
        \toprule
        Max speed $v_{max}$ ($km/h$) &\multicolumn{2}{c|}{30} &\multicolumn{2}{c|}{10} &\multicolumn{2}{c|}{30} &\multicolumn{2}{c|}{5} &\multicolumn{2}{c}{45}\\
        Max steering $s_{max}$ ($degree$) &\multicolumn{2}{c|}{40} &\multicolumn{2}{c|}{40} &\multicolumn{2}{c|}{10} &\multicolumn{2}{c|}{5} &\multicolumn{2}{c}{45}\\
        Wheel friction $\mu$ &\multicolumn{2}{c|}{0.7} &\multicolumn{2}{c|}{0.7} &\multicolumn{2}{c|}{0.7} &\multicolumn{2}{c|}{0.7} &\multicolumn{2}{c}{0.7}\\
        Engine force $F$ ($N$) &\multicolumn{2}{c|}{$350\sim550$} &\multicolumn{2}{c|}{$350\sim550$} &\multicolumn{2}{c|}{$350\sim550$} &\multicolumn{2}{c|}{$100\sim200$} &\multicolumn{2}{c}{$450\sim650$}\\
        Brake force $B$ ($N$) &\multicolumn{2}{c|}{$35\sim 80$} &\multicolumn{2}{c|}{$35\sim 80$} &\multicolumn{2}{c|}{$35\sim 80$} &\multicolumn{2}{c|}{$35\sim 80$} &\multicolumn{2}{c}{$35\sim 80$}\\
        \midrule
        SR$\uparrow$ (S | X) &47\% & 41\% &\best{62\%} & \second{56\%} &17\% & 22\% &\worstS{13\%} & \worstX{16\%} &36\% & 31\% \\
        SPL$\uparrow$ (S | X) &0.46 & 0.38 &\best{0.61} & \second{0.53}  &0.15 & 0.20 &\worstS{0.11} & \worstX{0.14} &0.33 & 0.28 \\
        Cost$\downarrow$ (S | X) &0.32 & 1.50 &\best{0.28} & \worstX{1.64} &\worstS{0.35} & 1.21 &0.30 & \second{1.10} &0.34 & 1.43 \\
        \bottomrule
    \end{tabular}
}
\end{center}
\end{table}

\vspace{-0.12in}

In this experiment, we evaluate the influence of mechanical structures on policy learning in a static obstacle avoidance task. We follow the setting of PointNav experiments in Section~\ref{tab:benchmark}. We perform PPO model training on three mobile machines, having significant disparities in max speed, max steering, wheel friction, engine force, and brake force. In addition, apart from comparing different machines, we compare different variations of one machine, \textit{i.e.}, COCO (base), COCO (mod-1), and COCO (mod-2), to find a better specification that can satisfy the demands of different use cases.
We evaluate all five models in straight street blocks (mainly static obstacles) and intersection street blocks (dense interactions with pedestrians) to test their behaviors in different scenarios, which are noted as ``S'' and ``X'' in Table~\ref{tab:cross_machine_eva}.
We use ``Success Rate (SR)'' and ``Success weighted by Path Length (SPL)'' to measure the mobility of agents, while use ``Cost'' to measure the safety of agents.

Results are shown in Table~\ref{tab:cross_machine_eva}.
The wheelchair has the most \textit{conservative design}, with the lowest max speed, max steering angle, and engine force. It achieves the best performance in Cost at the intersection scenario, which indicate the conservative design significantly improve its safety when navigating through the crowd. However, it has the worst performance in both straight and intersection street blocks, which indicates conservative design will influence the mobility of a machine to some degree.
The mobility scooter has the most \textit{aggressive design}, with the highest max speed, max steering angle, and engine force. It can achieve better performance than the wheelchair and COCO (mod-2). However, it has a larger Cost than the wheelchair and COCO (mod-2) on average, especially in the intersection scenario, which indicates a degradation of safety ability caused by its aggressive design.
The COCO (base) and its variations have a \textit{medium design} between the wheelchair and mobility scooter. Comparing COCO (base) with its variations, we can observe that based on the raw design, decreasing its max speed (mod-1) to 10$km/h$ can significantly improve its performance in both mobility and safety, while decreasing its max steering angle will diminish its performance remarkably.
These results could assist designers in finding improved mechanical structures for mobile machines to meet various application scenarios.

\vspace{-0.15in}
\subsection{Ablation Study}
\label{sec:ablation_study}
\vspace{-0.1in}

In this section, we evaluate the generalizability, scaling ability, and effects of the density of static objects and dynamic agents.
For unified evaluations, we use PPO for all ablation studies. Except for the results on dynamic density, we use the PointNav task. Observations and hyperparameters remain the same for model training across different evaluations.

\begin{figure*}[h!]
    \centering
    \vspace{-0.2in}    
    \includegraphics[width=1\linewidth]{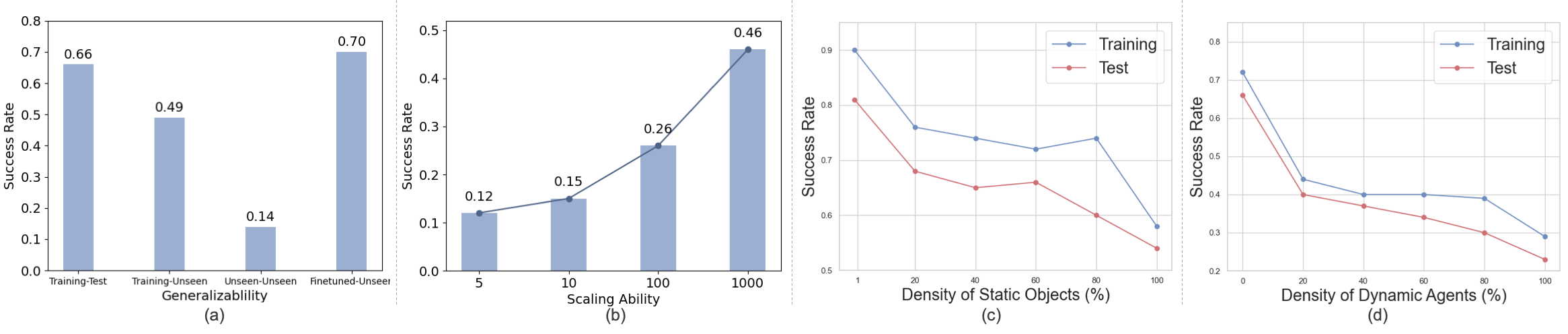}
    \vspace{-0.16in}
    \caption{\textbf{Ablation study.} (a) Evaluation of generalizability. (b) Evaluation of scaling ability. (c) Evaluation of the density of static objects. (d) Evaluation of the density of dynamic agents.}
    \vspace{-0.12in}
    \label{fig:ablation}
\end{figure*}

\noindent
\textbf{Evaluation of generalizability.}
To evaluate the generalizable ability of agents trained on data generated by MetaUrban, we compare the success rate of four settings in Figure~\ref{fig:ablation} (a).
Setting-1 and Setting-2 are the results of training on MetaUrban-train while testing on MetaUrban-test and MetaUrban-unseen, respectively. We can observe a performance drop on MetaUrban-unseen. However, the unseen evaluation results still achieve $49\%$ success rate facing various out-of-distribution scenes, demonstrating the strong generalizability of models trained on large-scale data created by MetaUrban.
Setting-3 and Setting-4 are the results of direct training on MetaUrban-finetune, and fine-tuning on MetaUrban-finetune from the pre-trained model on MetaUrban-train.
Compared between Setting-2 and Setting-3, we can observe an obvious performance drop, which is caused by an underfitting of the insufficient and complex fine-tuning data. Setting-4 outperforms Setting-3 by a large margin, demonstrating that the model trained on the MetaUrban-12K dataset can provide informative priors as good initializations for quick tuning.

\noindent
\textbf{Evaluation of scaling ability.}
To evaluate the scaling ability of MetaUrban's compositional architecture, we train models on a different number of generated scenes, from 5 to 1,000. As shown in Figure~\ref{fig:ablation} (b), the performance improves remarkably from 12$\%$ to 46$\%$, as we include more scenes for training, demonstrating the strong scaling ability of MetaUrban.
MetaUrban's compositional nature has the potential to extend more diverse scenes with a larger element repository in the future, which could further boost the agent's performance.

\noindent
\textbf{Evaluation of static and dynamic density.}
To evaluate the influence of static object density and dynamic environmental agents, we evaluate the different proportions of them on the PointNav and SocialNav tasks, respectively, from 1$\%$ to 100$\%$.
As shown in Figure~\ref{fig:ablation} (c) and (d), with the increasing density of both static objects and dynamic agents, the success rates of both train and test experience dramatic degradations, demonstrating the challenges for embodied agents when facing crowded scenes. In our experiments, we observe many interesting failure cases that can indicate promising future directions to improve AI’s performance. We showcase some videos on the project page.

\vspace{-0.15in}
\section{Conclusion}
\vspace{-0.15in}

We propose a new compositional simulator, MetaUrban, to facilitate embodied AI and micromobility research in urban scenes. MetaUrban can generate infinite interactive urban environments with complex scenes, diverse obstacles, and diverse movements of pedestrians and other mobile agents. These environments used as training data can significantly improve the generalizability and safety of the embodied AI underlying mobile machines. We commit ourselves to developing the open-source simulator and fostering community efforts to turn it into a sustainable infrastructure.

\bibliography{iclr2025_conference}
\bibliographystyle{iclr2025_conference}

\clearpage

\appendix

\clearpage
{
\section*{Appendix}
In the appendix, we present more details of MetaUrban. In Section~\ref{sec:metaurban_visualization}, we illustrate samples of static and dynamic scenes, as well as static and dynamic assets in MetaUrban. In Section~\ref{sec:metaurban_simulator}, we elaborate on the technical details of the MetaUrban simulator. In Section~\ref{sec:metaurban_dataset}, we will give the construction details and statistics of the MetaUrban-12K dataset. In Section~\ref{sec:exp_detail}, we will present details of implementations in our benchmarks and further exploration in cross-machine evaluations. In Section~\ref{sec:unique_challenges}, we delve into and validate unique challenges in urban micromobility. In Section~\ref{sec:data_sheet}, we provide the datasheet of MetaUrban and MetaUrban-12K dataset. In Section~\ref{sec:performance}, and ~\ref{sec:robustness}, we evaluate the performance and robustness of models trained on MetaUrban. In Section~\ref{sec:discussion}, we discuss the impacts, limitations, sim-to-real generalization, and future work of MetaUrban.

\startcontents[sections]
\printcontents[sections]{l}{1}{\setcounter{tocdepth}{3}}
}
\clearpage

\section{MetaUrban Visualization}
\label{sec:metaurban_visualization}

\subsection{Static Scene Samples}

\begin{figure*}[h!]
    \centering
    \includegraphics[width=1\linewidth]{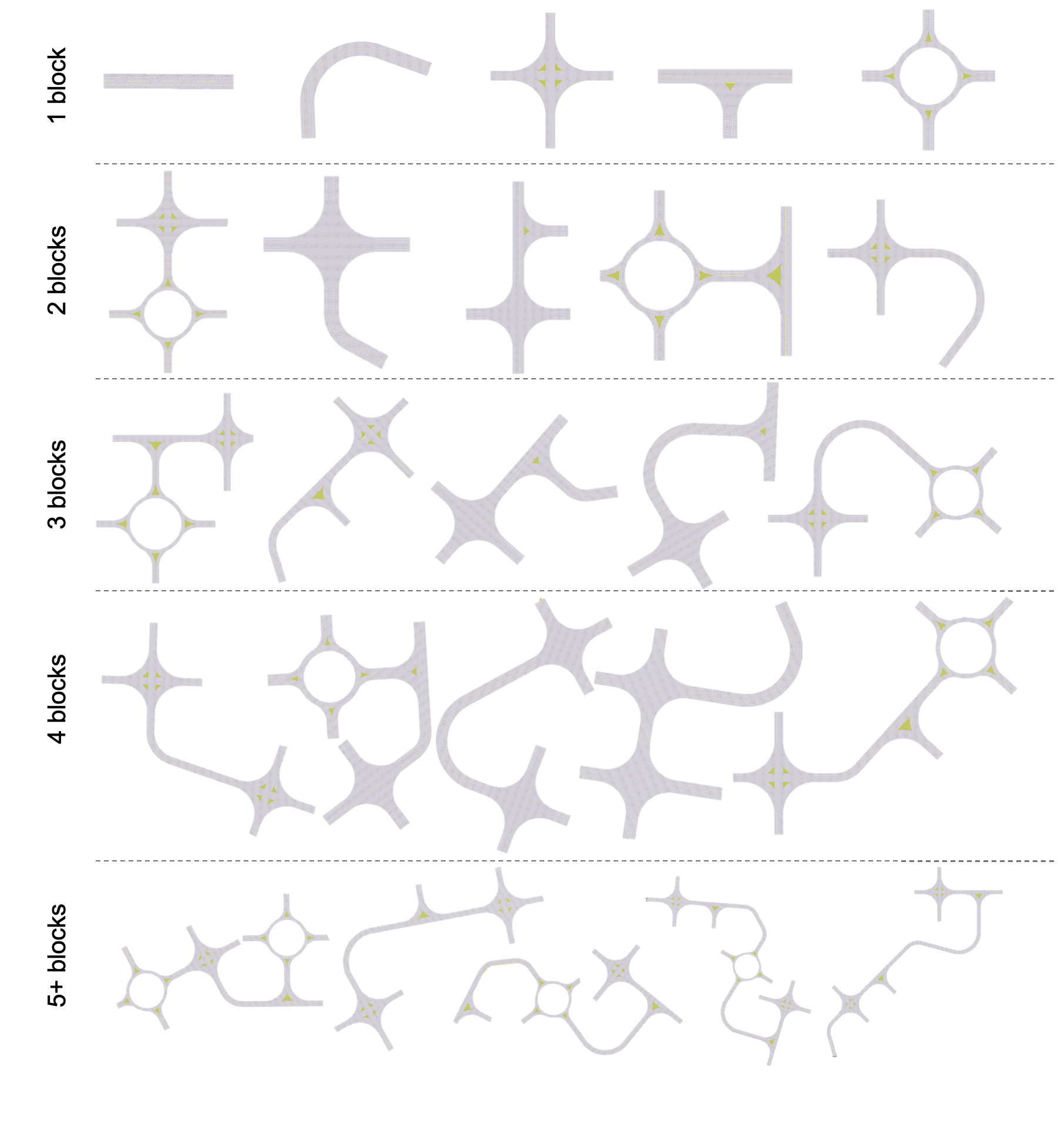}
    \caption{\textbf{Examples of block maps.} Generated block maps with a different number of street blocks.
    }
    \label{fig:street_blocks}
\end{figure*}

\noindent
\textbf{Street blocks.}
We design five typical street block categories -- straight, curve, intersection, T-junction, and roundabout. In the simulator, to form a large map with several blocks, we can sample the category, number, order, lane number, and other related parameters of the blocks. We use the algorithm Block Incremental Generation (BIG) proposed in MetaDrive~\citep{li2022metadrive} to generate the target road network defined by users. Figure~\ref{fig:street_blocks} provides demonstrations of generated street maps composed of different numbers of blocks.

\noindent
\textbf{Ground layouts.}
We construct seven typical templates for sidewalks, more details about the design and the generation process are given in the Section~\ref{sec:layoutgeneration}.

\begin{figure*}[h!]
    \centering
    \includegraphics[width=1\linewidth]{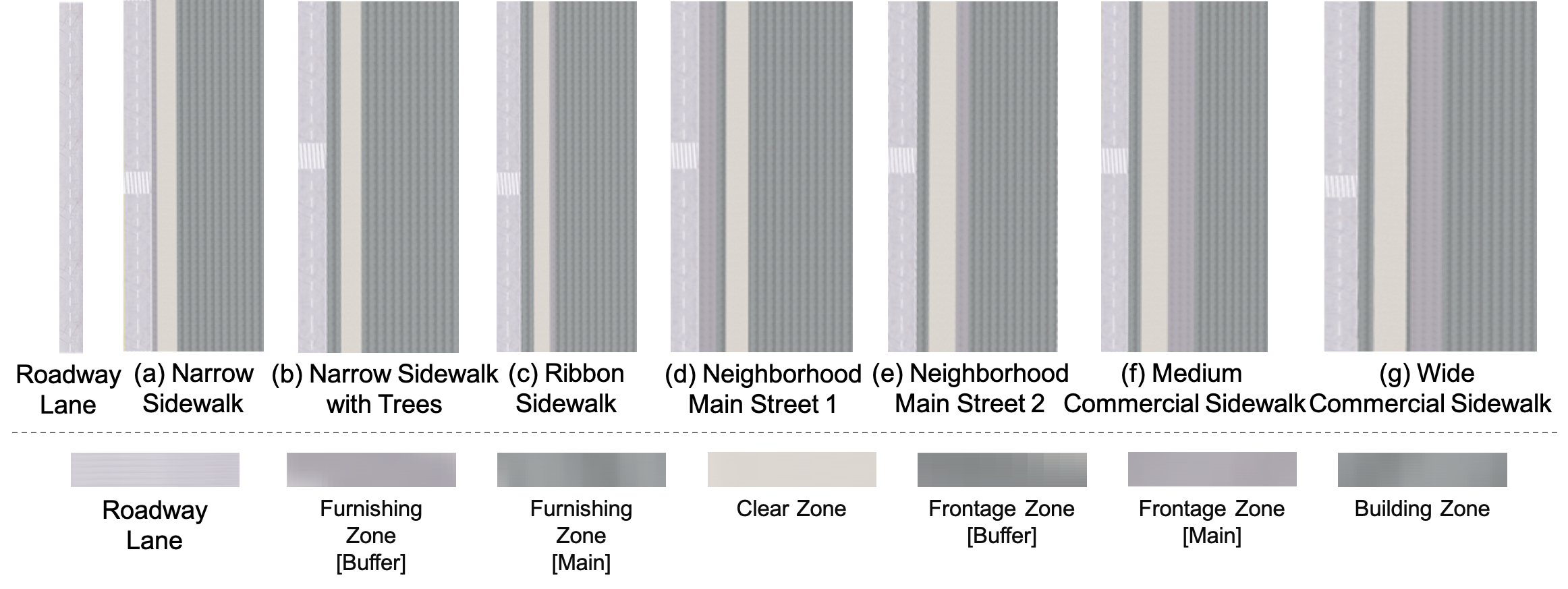}
    \caption{\textbf{Examples of sidewalks.} Generated sidewalks with seven templates (a) to (g).
    }
    \label{fig:sidewalk_example}
\end{figure*}

\begin{figure*}[h!]
    \centering
    \includegraphics[width=1\linewidth]{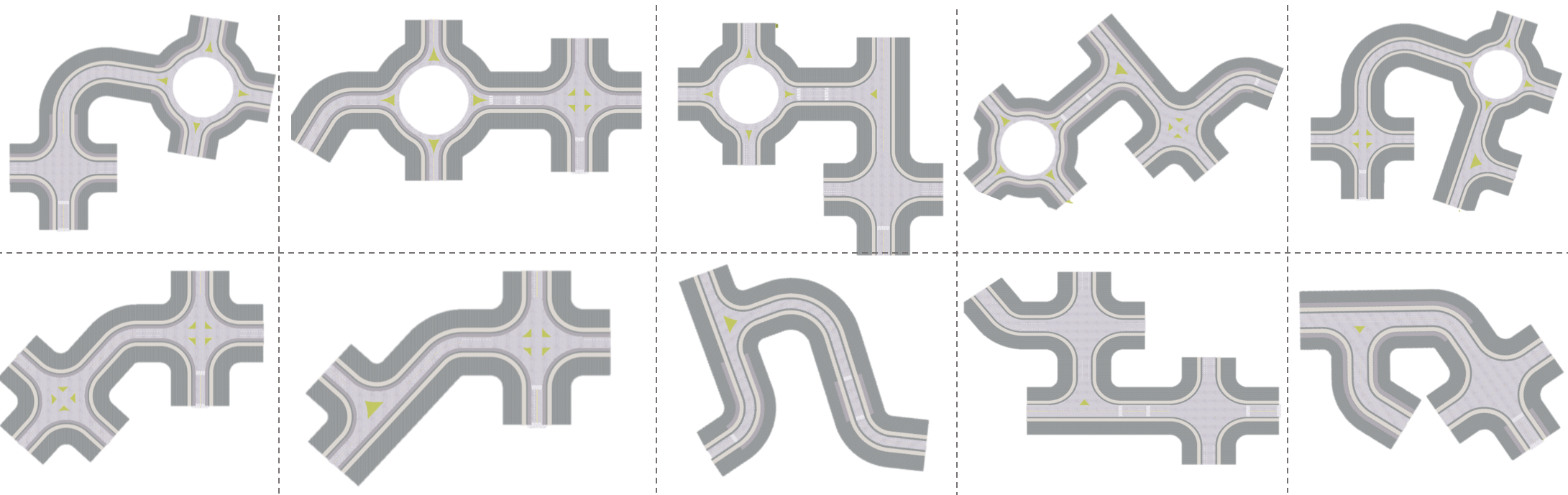}
    \caption{\textbf{Examples of block maps with sidewalks.} Generated block maps with a different type of sidewalks.
    }
    \label{fig:samples_side}
\end{figure*}

As shown in Figure~\ref{fig:sidewalk_example}, different types of sidewalks can be sampled on the same street block; each type has its unique division and specification of functional zones. Figure~\ref{fig:samples_side} further shows several block maps with a different type of sidewalks.

\clearpage

\begin{figure*}[h!]
    \centering
    \includegraphics[width=1\linewidth]{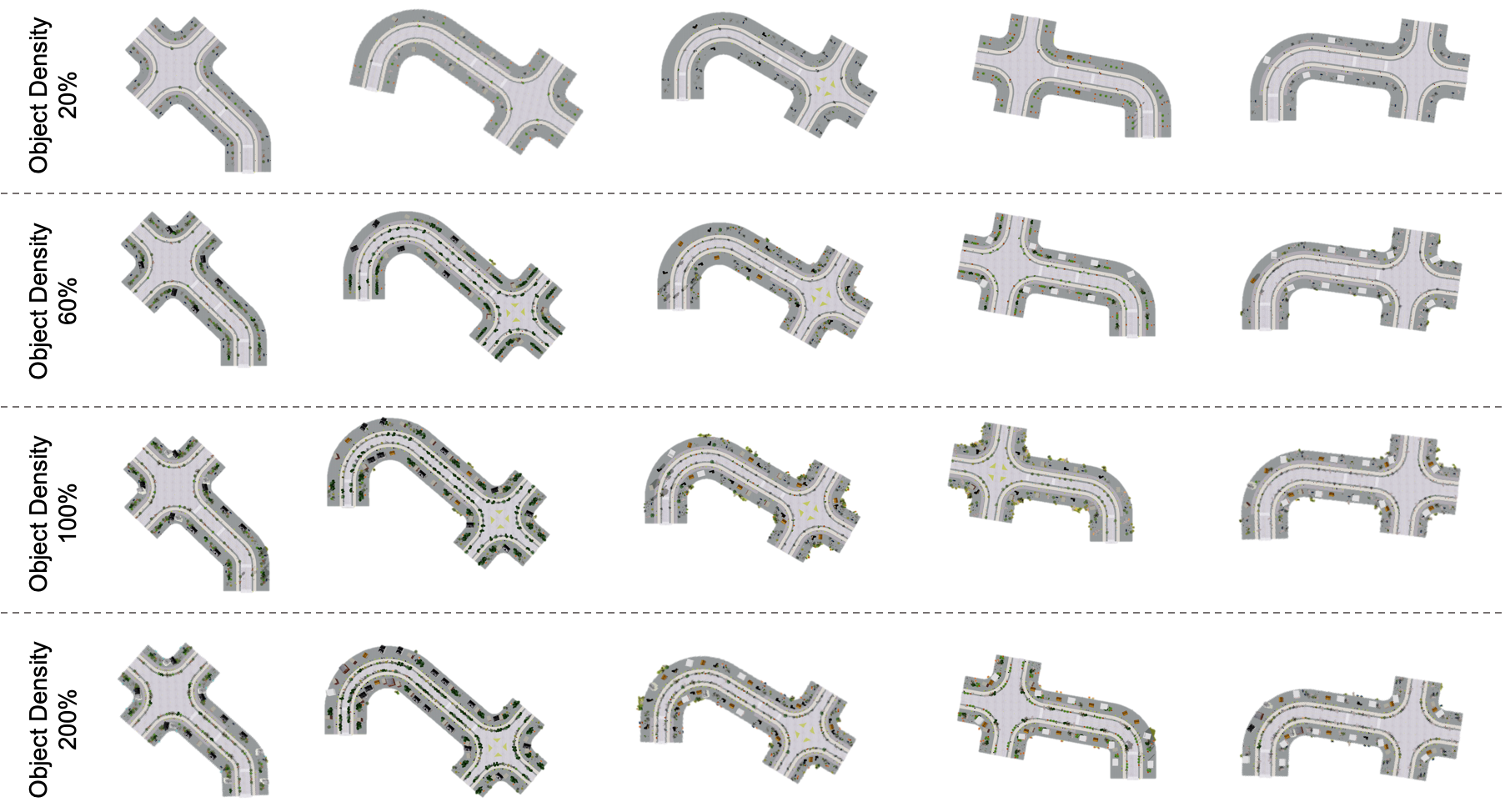}
    \caption{\textbf{Examples of block maps with different object densities.} Each row is 5 randomly sampled block maps with one object density, from 20$\%$ to 200$\%$.
    }
    \label{fig:samples_obj}
\end{figure*}

\begin{figure*}[h!]
    \centering
    \includegraphics[width=1\linewidth]{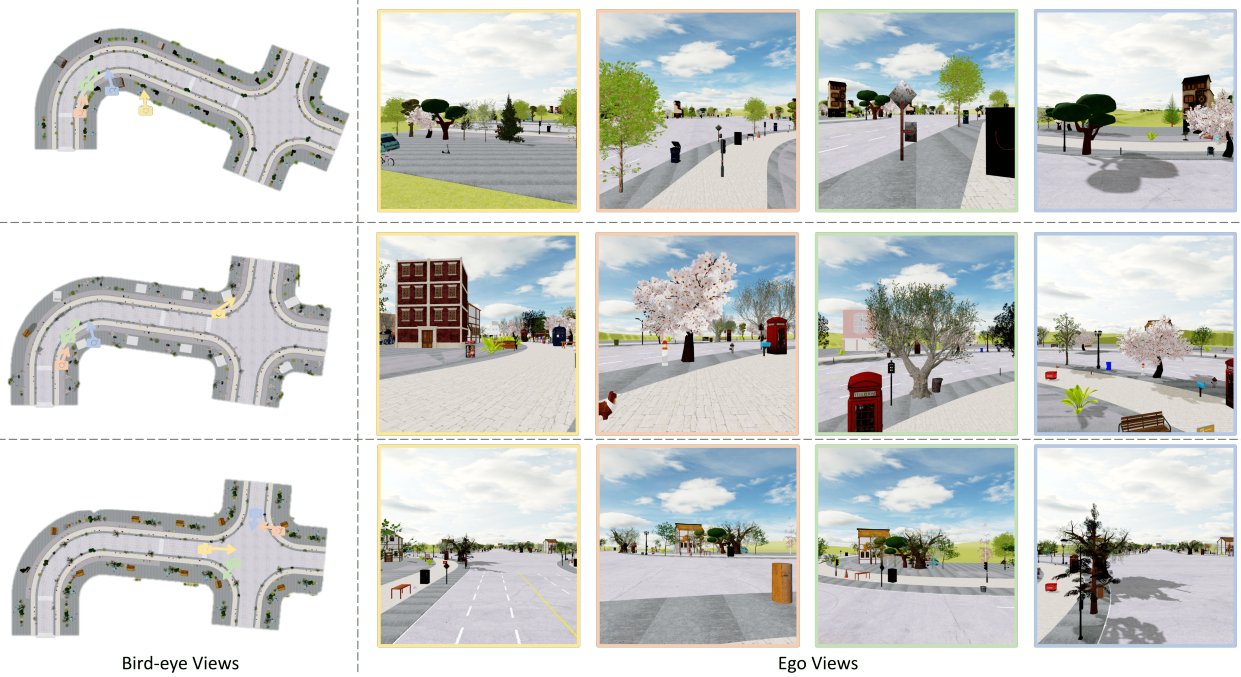}
    \caption{\textbf{Examples of ego-view results in static scenes.} Each row is a different object placement with the same object density (60\%). For each row, we sample 4 viewpoints to show ego-view results.
    }
    \label{fig:obj_random}
\end{figure*}

\noindent
\textbf{Static objects.}
To generate static objects, we build the object placement distribution conditioned on geometric zones of sidewalks, which will be discussed in Section~\ref{sec:layoutgeneration}. To better distinguish between the difficulty of scenes on the same road network, we use the object density $\rho_{s}$ to control the crowding level on the sidewalk. This indicates the ratio of the minimum distance to the default distance between objects. Figure~\ref{fig:samples_obj} shows block maps with different object densities. We can observe that when the density increases, the walkable region will become more and more crowded. Figure~\ref{fig:obj_random} further gives ego-view results by randomly sampling viewpoints on block maps.

\clearpage

\subsection{Dynamic Scene Samples}

\begin{figure*}[h!]
    \centering
    \includegraphics[width=1\linewidth]{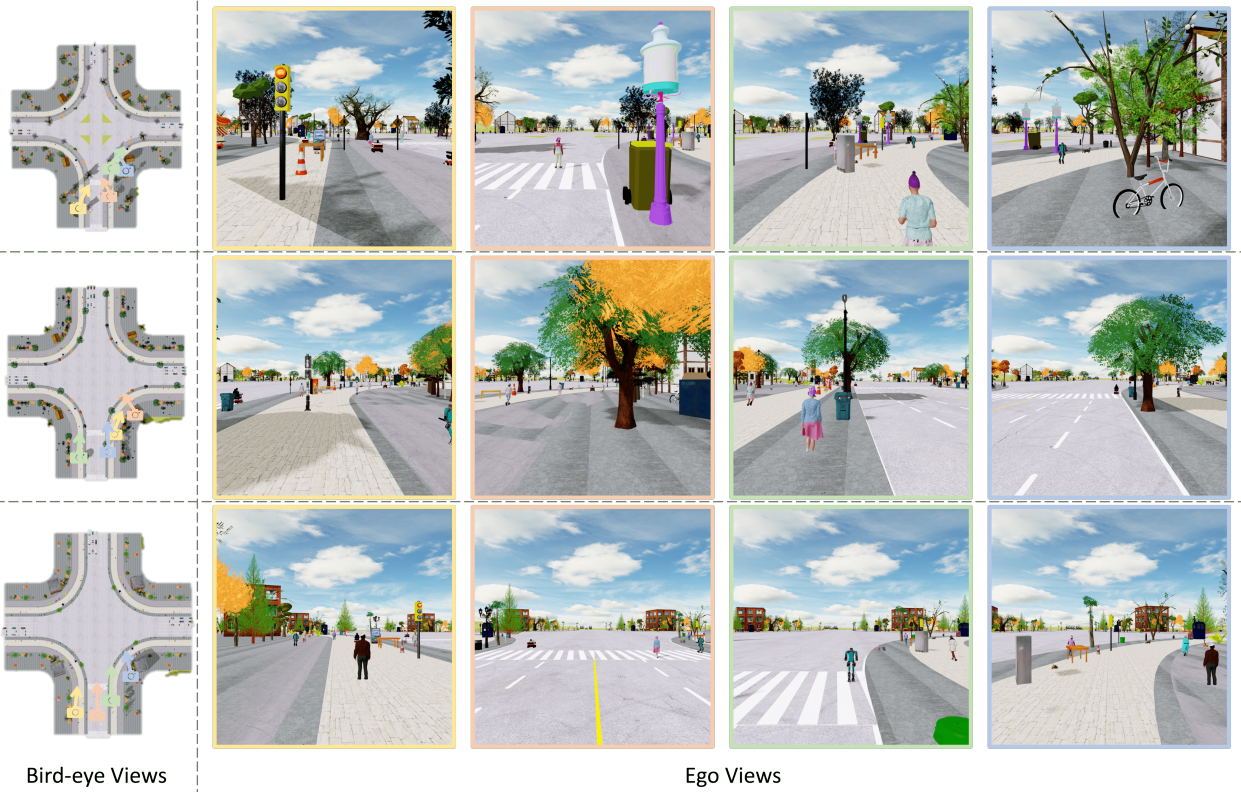}
    \caption{\textbf{Examples of ego-view results in dynamic scenes.} Each row is a different specification of dynamics (appearances, movements, and trajectories) with the same dynamic density (100\%). For each row, we sample 4 viewpoints to show ego-view results.
    }
    \label{fig:dynamic_random}
\end{figure*}

Dynamic agents such as pedestrians, vulnerable road
users like bikers (skateboarders, scooter riders), mobile machines (delivery bots, electric wheelchairs, robot dogs, and humanoid robots), and vehicles will be present in the environment. The density of dynamic agents can be controlled with dynamic density ratio $\rho_{d}$. Figure~\ref{fig:dynamic_random} shows ego-view results by randomly sampling viewpoints on block maps. The urban spaces are well populated with different agents.

\subsection{Static Asset Samples}

\begin{figure*}[h!]
    \centering
    \includegraphics[width=0.9\linewidth]{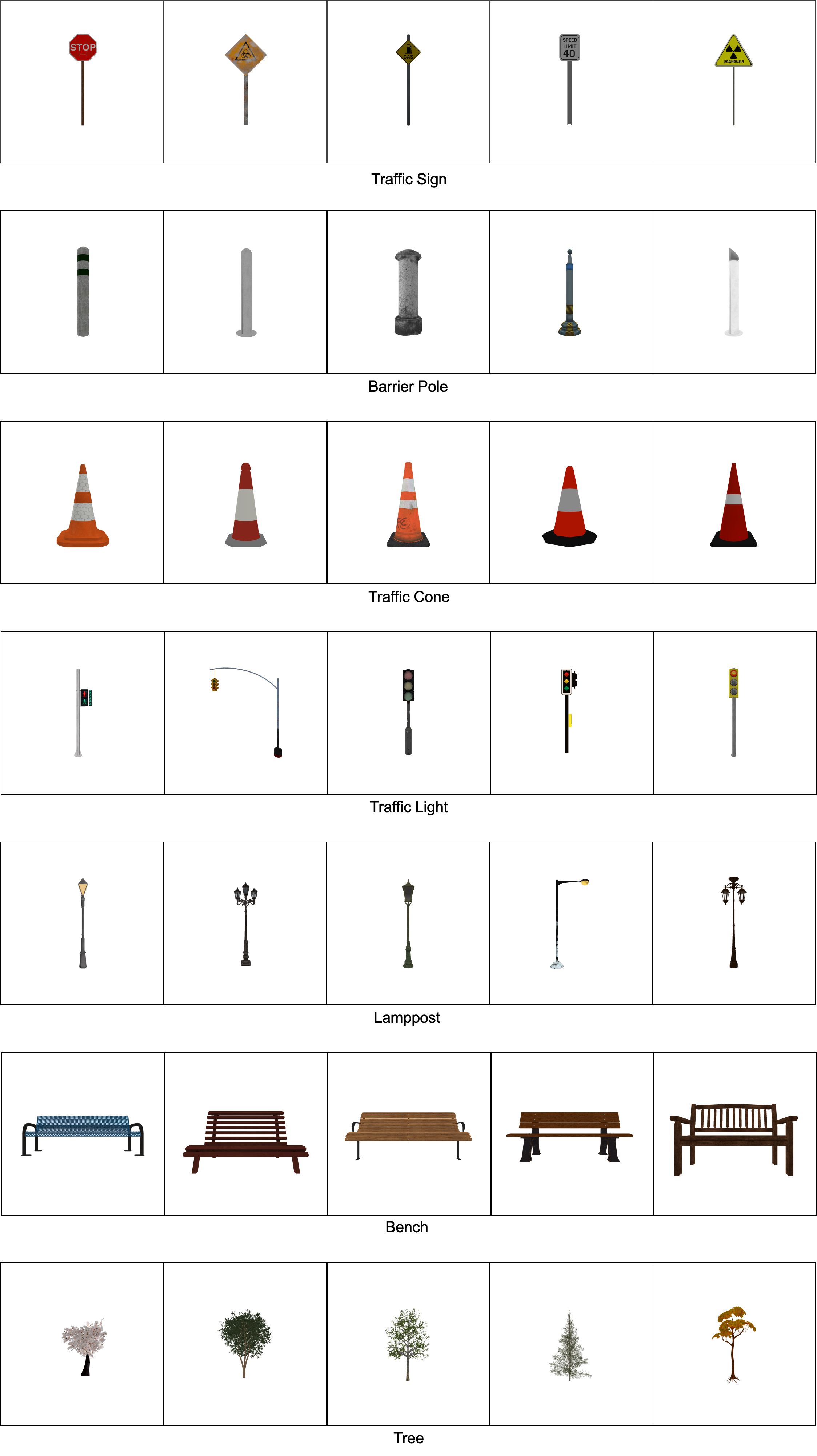}
    \caption{\textbf{Examples of static objects -- standard infrastructure.}
    }
    \label{fig:Object_Standard}
\end{figure*}

\begin{figure*}[h!]
    \centering
    \includegraphics[width=0.9\linewidth]{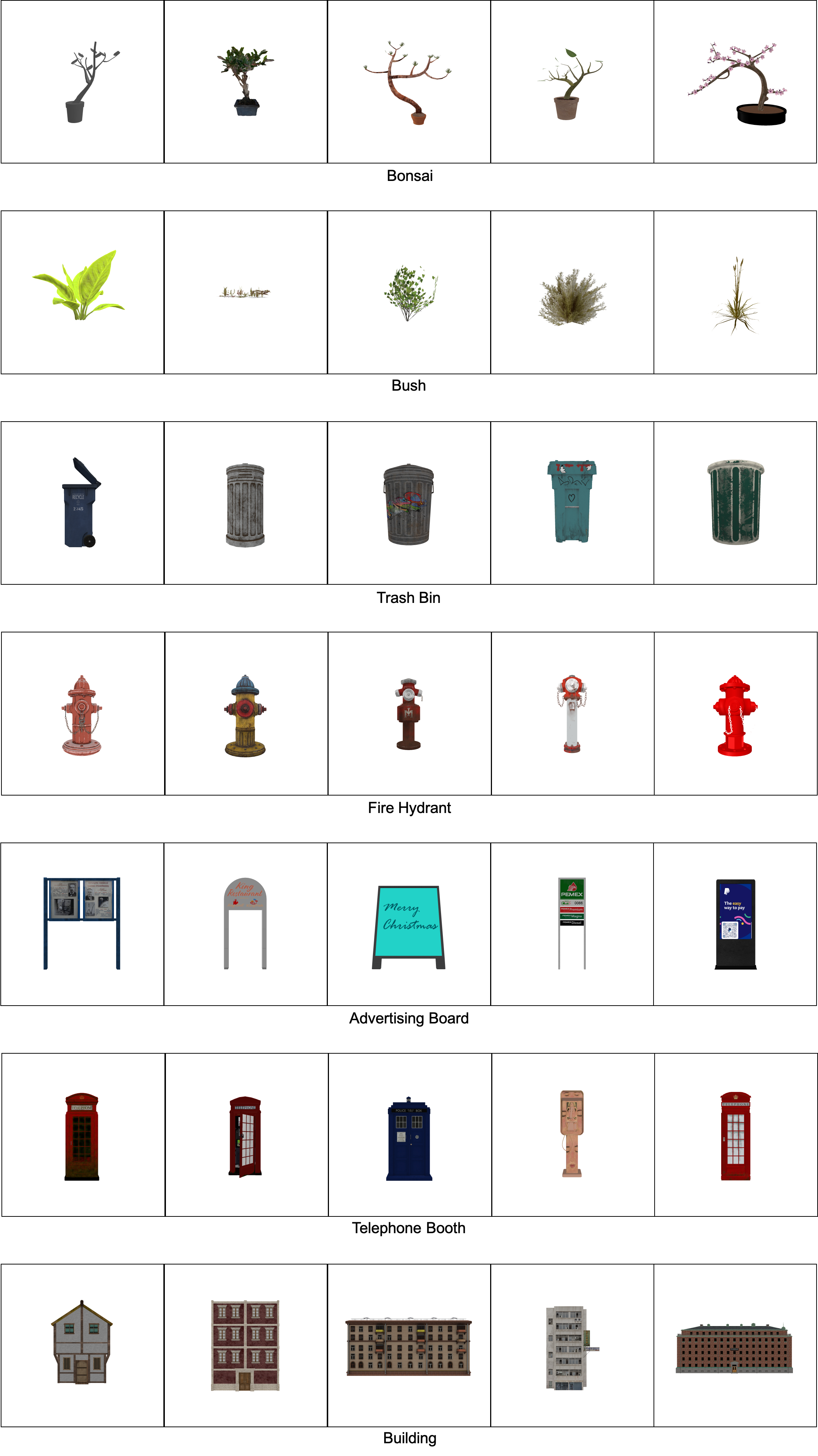}
    \caption{\textbf{Examples of static objects -- non-standard infrastructure.}
    }
    \label{fig:Object_Non_Standard}
\end{figure*}

\begin{figure*}[h!]
    \centering
    \includegraphics[width=0.9\linewidth]{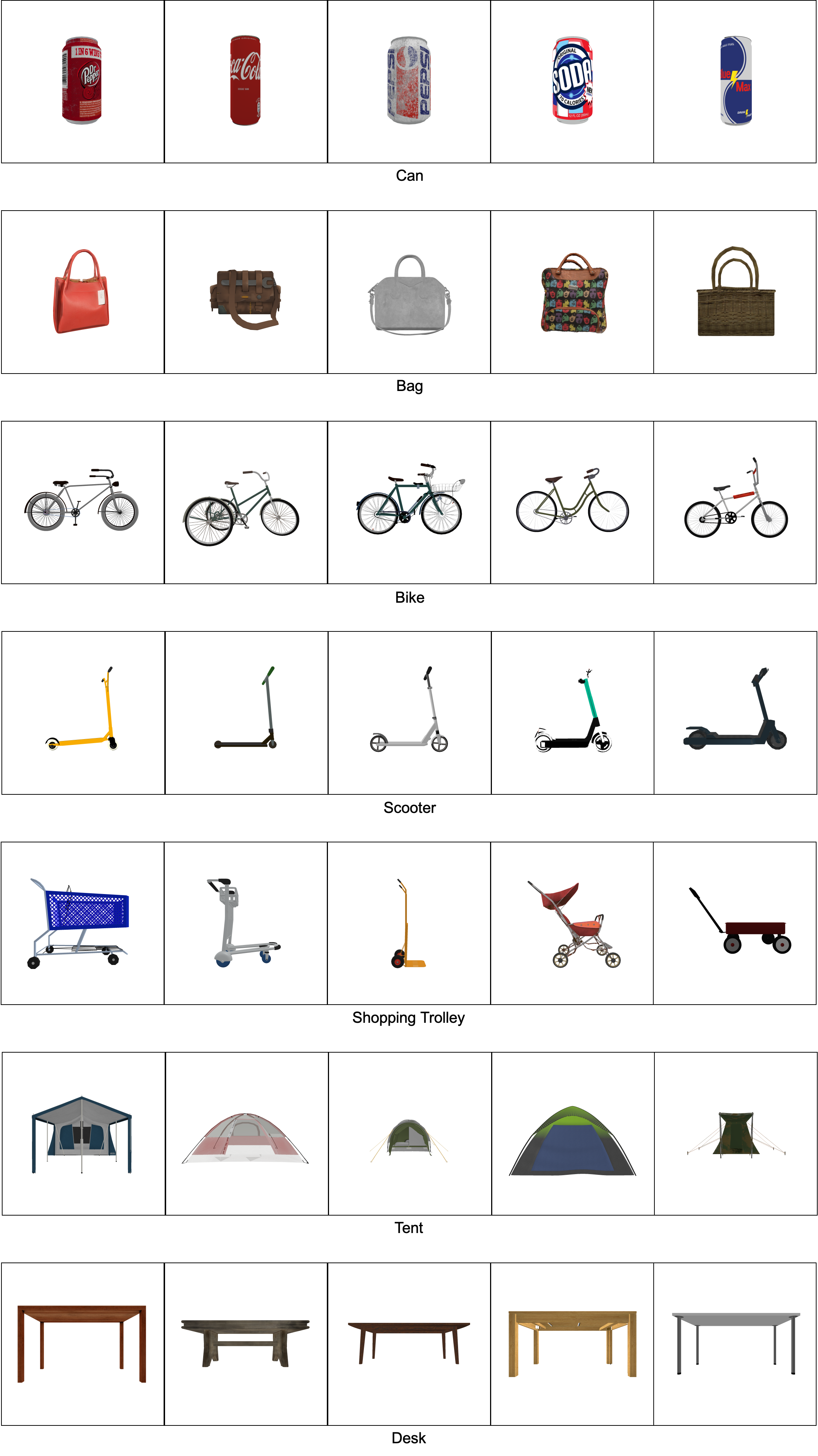}
    \caption{\textbf{Examples of static objects -- clutter.}
    }
    \label{fig:Object_clutter}
\end{figure*}

We provide 10,000 high-quality static object assets. The roadside infrastructure is divided into three categories: 
1) Standard infrastructure, including poles, trees, and signs, is placed at regular intervals along the road. 
2) Non-standard infrastructure, such as buildings, bonsai, and trash bins, is placed randomly within designated zones. 
3) Clutter, such as drink cans, bags, and bicycles, is scattered randomly across all functional zones.
Figure~\ref{fig:Object_Standard} \ref{fig:Object_Non_Standard} and \ref{fig:Object_clutter} show examples of these three categories respectively.

\clearpage

\subsection{Dynamic Asset Samples}

\begin{figure*}[h!]
    \centering
    \includegraphics[width=0.9\linewidth]{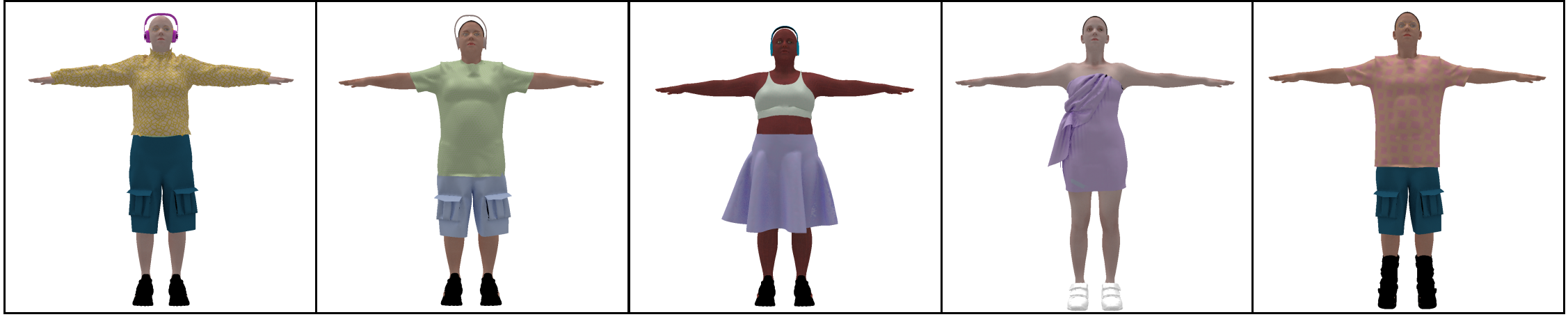}
    \caption{\textbf{Examples of dynamics -- rigged humans.}
    }
    \label{fig:human_asset}
\end{figure*}

\noindent
\textbf{Human assets.}
MetaUrban provides 1,100 rigged 3D human models, sampled from 68 garments, 32 hairs, 13 beards, 46 accessories, and 1,038 cloth and skin textures from SynBody~\cite{yang2023synbody} dataset. Figure~\ref{fig:human_asset} shows randomly sampled humans, which have large variations.

\begin{figure*}[h!]
    \centering
    \includegraphics[width=0.9\linewidth]{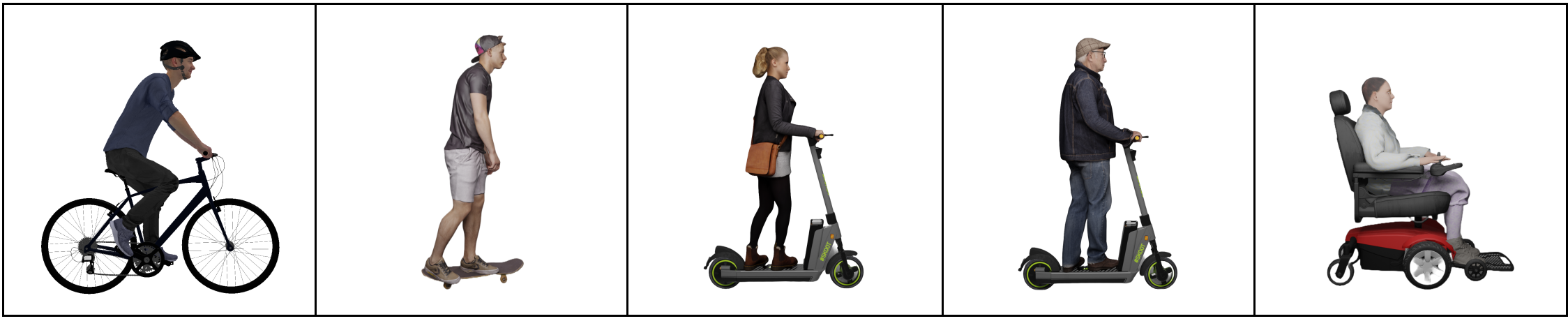}
    \caption{\textbf{Examples of dynamics -- vulnerable road users.}
    }
    \label{fig:vulnerable_road_user_asset}
\end{figure*}

\noindent
\textbf{Vulnerable road user assets.}
MetaUrban provides 5 kinds of vulnerable road users to form safe-critical scenarios. They are bikers, skateboarders, scooter riders, and electric wheelchair users, as shown in the first row of Figure~\ref{fig:vulnerable_road_user_asset}.
Note that electric wheelchairs, as a human-AI shared control system, can also be seen as mobile machines, not only vulnerable road users.

\begin{figure*}[h!]
    \centering
    \includegraphics[width=0.9\linewidth]{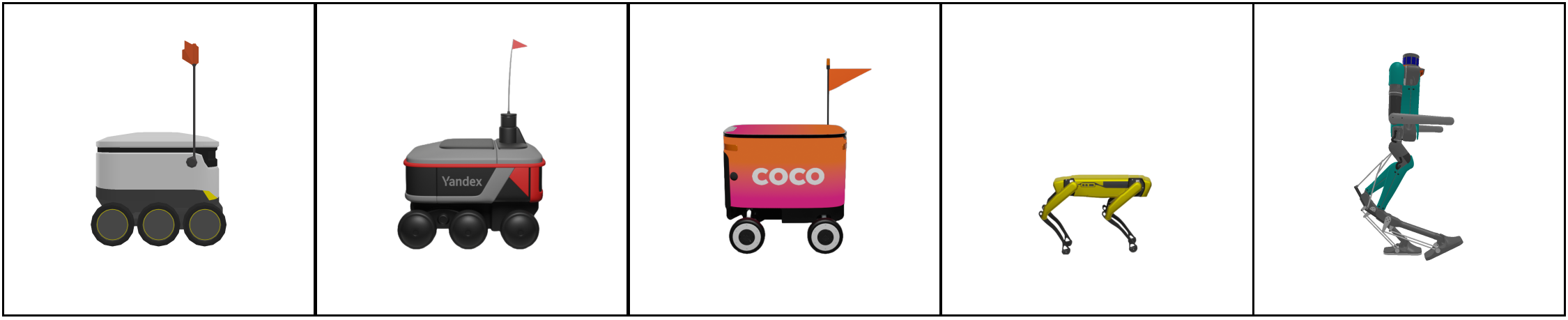}
    \caption{\textbf{Examples of dynamics -- mobile machines.}
    }
    \label{fig:robot_asset}
\end{figure*}

\noindent
\textbf{Mobile machine assets.}
MetaUrban provides 6 kinds of mobile machines: Starship, Yandex Rover, and COCO Robotics’ delivery bots, Boston Dynamic’s robot dog, Agility Robotics’ humanoid robot, and Drive Medical’s electric wheelchair. Figure~\ref{fig:robot_asset} shows the first 5 assets, while the electric wheelchair, as a cross-category asset (vulnerable road user and mobile machine), is shown in Figure~\ref{fig:vulnerable_road_user_asset}.

\begin{figure*}[h!]
    \centering
    \includegraphics[width=0.9\linewidth]{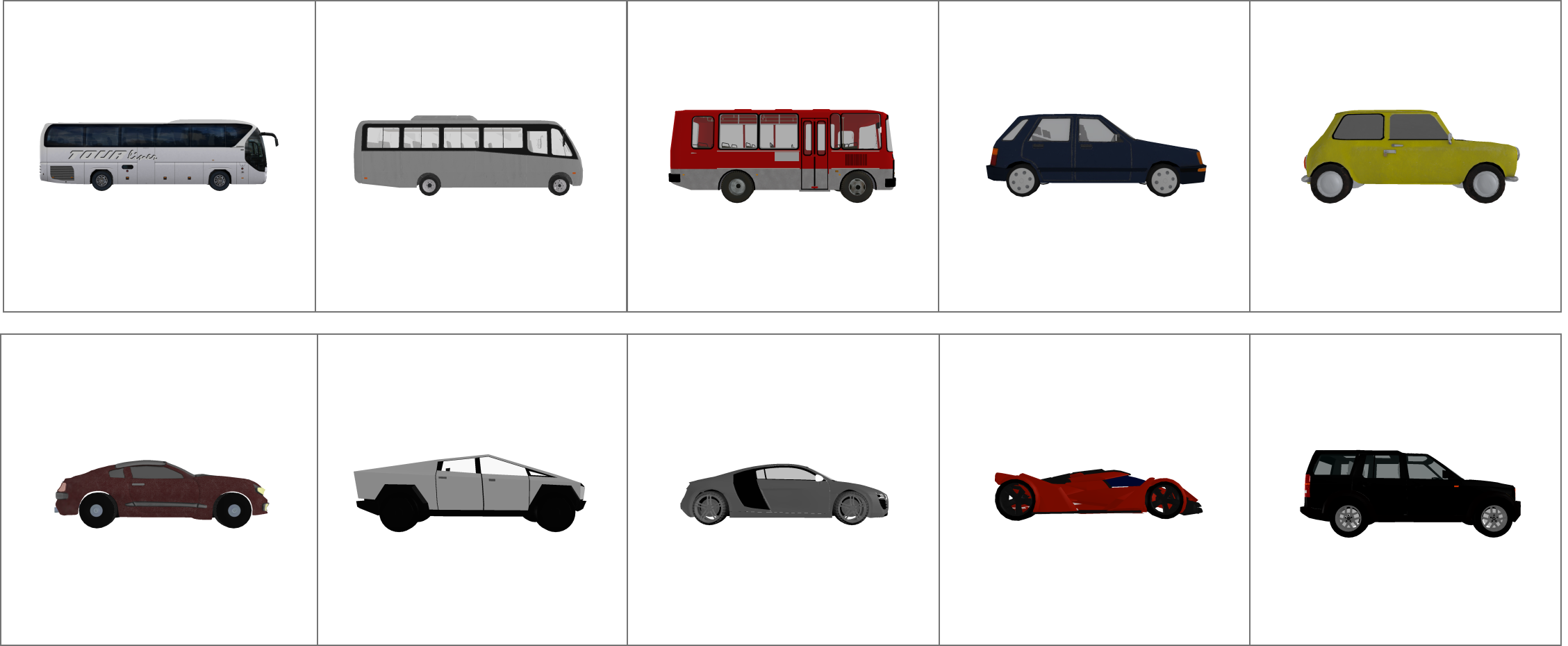}
    \caption{\textbf{Examples of dynamics -- vehicles.}
    }
    \label{fig:Object_Cars}
\end{figure*}

\textbf{Vehicle assets.}
MetaUrban provides 37 kinds of vehicles, covering different body types, sizes, and appearances. Figure~\ref{fig:Object_Cars} shows 10 sampled vehicles.

\section{MetaUrban Simulator}
\label{sec:metaurban_simulator}

\subsection{Layout Generation}
\label{sec:layoutgeneration}

This section gives details about the process we developed to procedurally generate scenes with sidewalks and crosswalks, as well as sample and place static objects on the sidewalk.

\begin{figure*}[h!]
    \centering
    \includegraphics[width=1\linewidth]{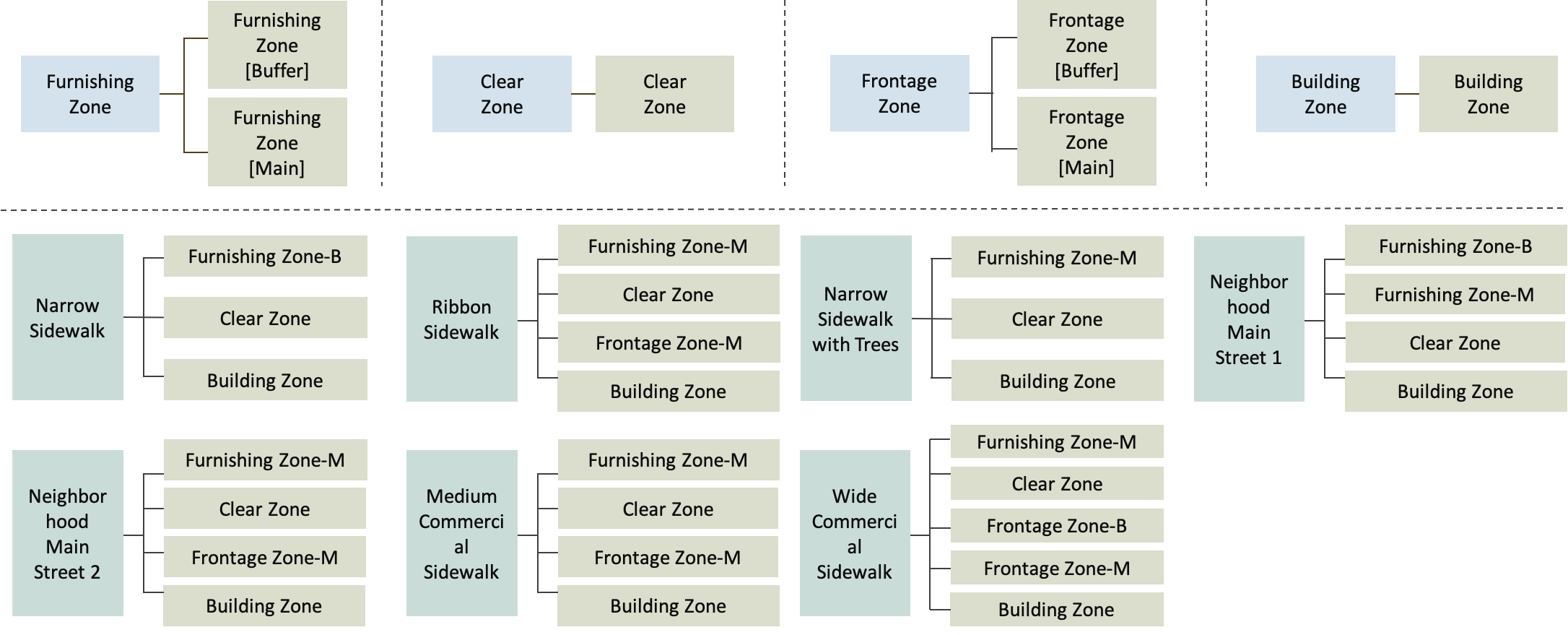}
    \caption{\textbf{Architecture of ground layouts.} (Top) The mapping from functional zones to geometric zones. (Bottom) Specifications of geometric zones for 7 sidewalk templates.
    }
    \label{fig:ground_plan2}
\end{figure*}

\noindent
\textbf{Ground plan.}
As shown in Figure~\ref{fig:ground_plan2} (Top), we define 4 functional zones \legendsquare{fun} and 6 geometric zones \legendsquare{geom} for sampling the type of sidewalks and choosing the distribution of parameters for each sidewalk component. 
As shown in Figure~\ref{fig:ground_plan2} (Bottom), we construct 7 typical templates for sidewalks; each type of them has its unique distribution of geometric zones. To match the distribution with the real world, we set the distribution of the zone width to a uniform distribution for each geometric zone; the maximum and minimum values of the uniform distribution are set according to the Global Street Design Guide~\citep{global2016global} provided by the Global Designing Cities Initiative.

To generate a scene, we will first sample the template of the sidewalk $z$ from its distribution $z\sim\mathcal{Z}_{\mathcal{T}}, \ \ \ \mathcal{Z}_{\mathcal{T}}=\{z_1,z_2,...,z_7\}$, followed by the sampling of widths of each geometric zone $w_i\sim f_w(z,i), \ \ \ \forall i\in \{1,2,...,7\}$, where $f_w(z,i)$ is the width distribution of the $i$th geometric zone under the sidewalk template $z$.

Crosswalks are crucial for the connectivity of scenes. MetaUrban provides candidates at the beginning and end of each roadway of a block. Then, locations of the crosswalk can be controlled by a crosswalk density parameter or be specified by users directly.

\begin{figure*}[h!]
    \centering
    \includegraphics[width=1\linewidth]{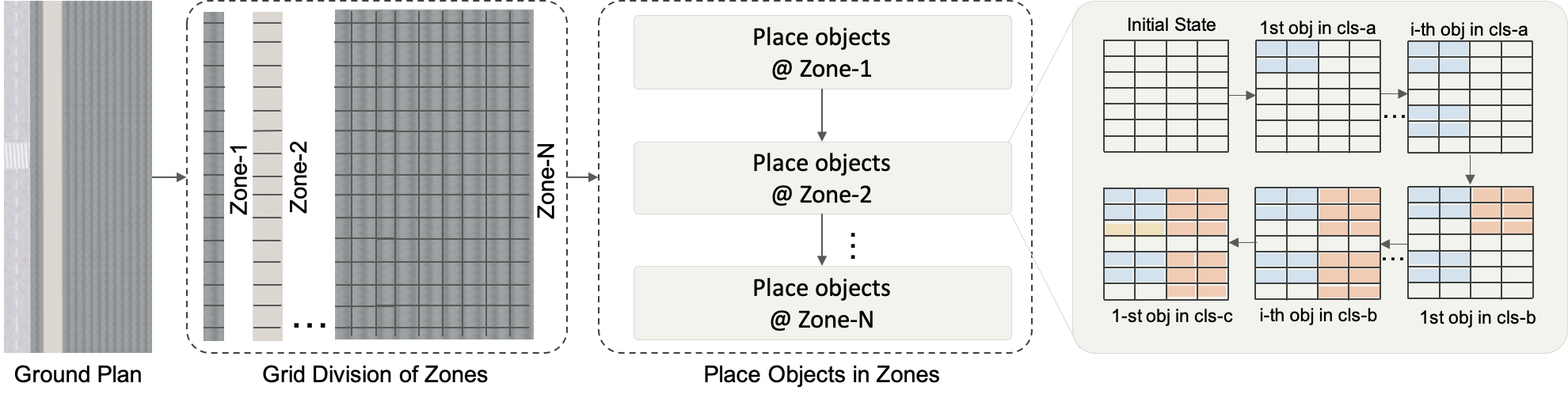}
    \caption{\textbf{Iterative object placement.}
    }
    \label{fig:place_obj}
\end{figure*}

\noindent
\textbf{Object placement.}
Figure~\ref{fig:place_obj} illustrates the iterative process of placing objects in the scene. First, we convert the polygon of the geometric zone of the sidewalk into rectangles. We will place objects on each functional zone or geometric zone independently. At each iteration of generating on the specific zone, we can obtain rectangles that are not occupied. Then we check from the starting region to the ending region for the current retrieved object class. We place it if possible, or we start to place the next class.
In the simulator, we use rectangle bounding boxes to represent all objects physically to adopt this object placement method.

\subsection{Object Retrieval}
\label{retrieval}

\noindent
\textbf{Distribution extraction.}
Distinguished from the recent indoor simulation platform, there are no ready-to-use high-quality asset datasets for urban spaces.
Urban spaces have their unique data distribution, such as the infrastructure built by the urban planning administration (``fire hydrants'' and ``bus stops'') and clutters placed by people (``scooters'' and ``advertising boards''). Thus, we design a real-world distribution extraction method to get a description pool depicting what objects are frequently shown in urban spaces.

We first leverage off-the-shelf scene understanding datasets -- Mapillary Vistas~\citep{neuhold2017mapillary} and CityScape~\citep{cordts2016cityscapes}. Using the provided annotation polygon, we find the overlapping object with the sidewalk and get a list of 90 objects that are with high frequency t1o be put in the urban space (such as ``tree'' and ``bench'').
However, the number of objects is limited because of the closed-set definitions in the image datasets. To get broader object distribution from the real world, we introduce two open-set sources -- worldwide Google Street data and urban planning description data.

For the Google Street data, we collect 25,000 urban space images from 50 countries across six continents. The selection of image locations was performed by randomly sampling points along the major roads of cities using OpenStreetMap’s \citep{OpenStreetMap} road network. Image orientation was determined based on road gradient to enhance the relevance of captured scenes.
For object detection in these images, we initially employed GPT-4o \citep{gpt-4o} to generate a list of candidate objects. This was followed by the application of Grounded-Dino \citep{liu2023grounding} to obtain bounding boxes for these objects. We refined these boxes using non-maximum suppression (NMS) to ensure the accuracy of object identification.

\begin{figure*}[h!]
    \centering
    \includegraphics[width=1\linewidth]{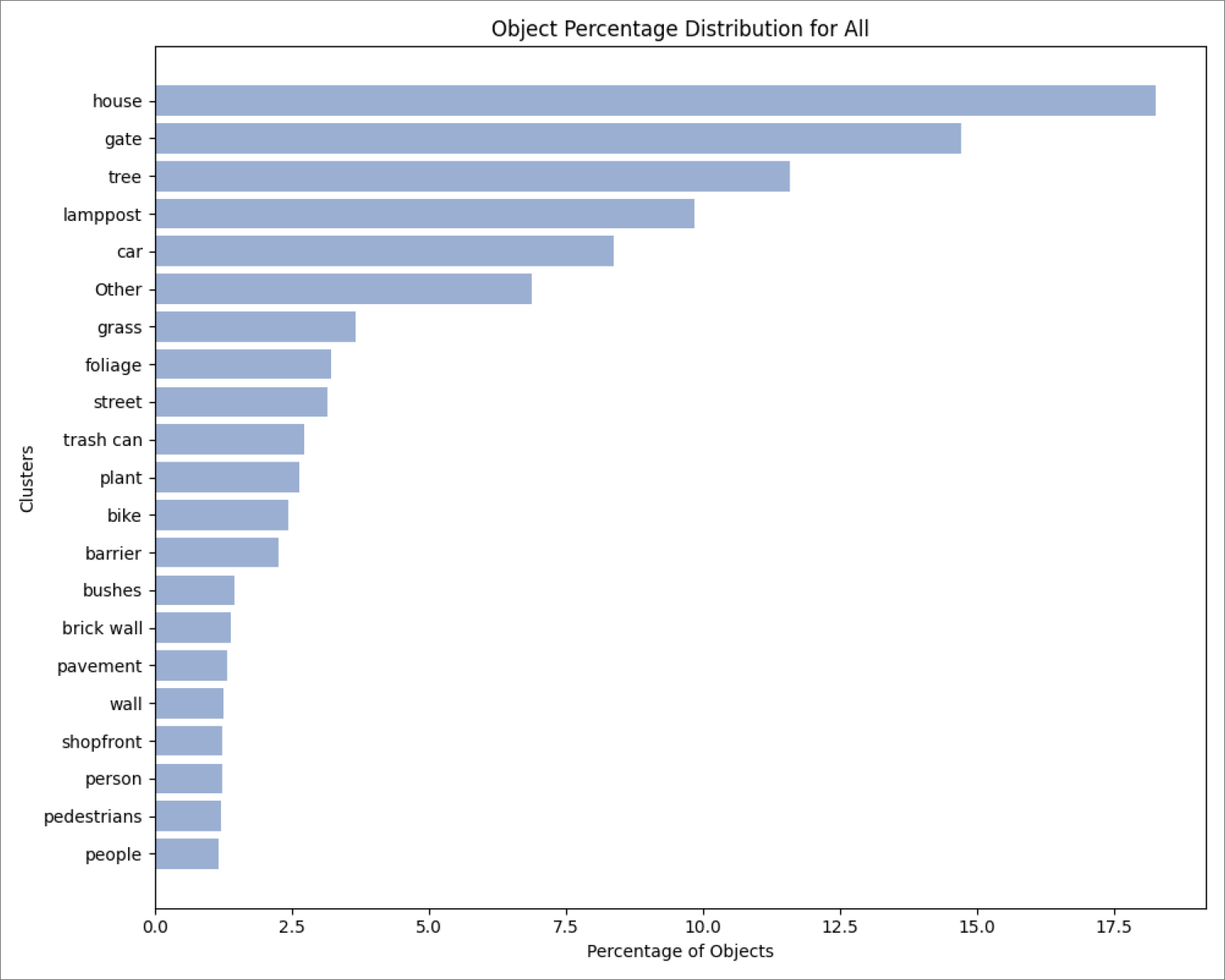}
    \caption{\textbf{Distribution of objects in urban spaces for all collected data worldwide.}
    }
    \label{fig:worldwide_distribution}
\end{figure*}

Further refinement was achieved through the use of the Grounded-SAM model~\citep{ren2024grounded}, an open-set segmentation approach, which filtered the bounding boxes to identify objects specifically located in public urban spaces. A key part of our method involves determining overlaps between identified objects and sidewalks. For each object detected, we calculate its spatial intersection with sidewalk regions derived from the datasets. This overlap analysis helps in curating a list of objects that are relevant to public urban spaces.

To address the diverse descriptions generated by GPT-4o \citep{gpt-4o} and ensure semantic uniformity, we cluster the embeddings of descriptions using DBSCAN \citep{ester1996density}, which result in 1,075 distinct object clusters with unique descriptors, such as "a gray trash bin" and "potted cactus".  We use ``all-mpnet-base-v2'' model from SentenceTransformers \citep{reimers-2019-sentence-bert} to embed each description.

For the urban planning description data, we get a list of 50 essential objects in public urban spaces (such as ``drinking fountains'' and ``bike racks'') through a thorough survey of urban design handbooks.
Finally, by combining these three data sources, we can get an object description pool with 1,215 items of descriptions that can form the real-world object category distribution.

Figure~\ref{fig:worldwide_distribution} illustrates the distribution of objects in urban space extracted from all of the worldwide collected data. Houses, gates, and trees emerge as universal elements, dominating the urban landscapes across all depicted countries, reflecting their fundamental role in both urban and rural settings.

\begin{figure*}[h!]
    \centering
    \includegraphics[width=1\linewidth]{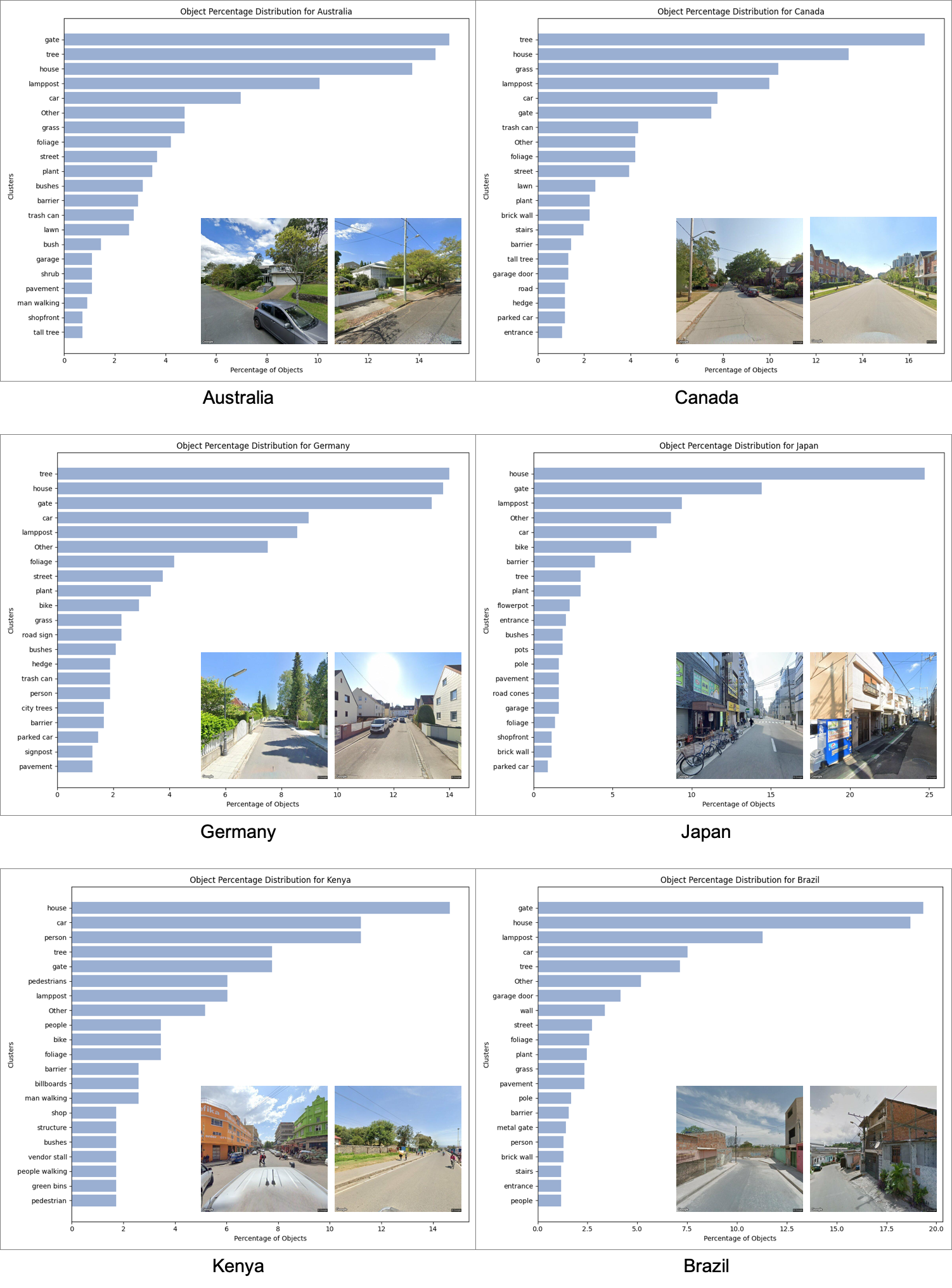}
    \caption{\textbf{Distribution of objects in urban spaces across different countries.} Two example images are shown together with each distribution figure, demonstrating large variations among different countries.
    }
    \label{fig:country_distribution}
\end{figure*}

Figure~\ref{fig:country_distribution} illustrates the object distribution of example countries from 6 continents, showcasing distinct environmental and cultural characteristics through object prevalence. The data also highlights notable regional distinctions: Japan, for instance, features a higher incidence of poles and road cones, hinting at unique aspects of its urban infrastructure. In contrast, Brazil's considerable frequency of gates and metal gates suggests prominent architectural and security preferences. Such variances not only reveal the diverse urban aesthetics and functional priorities across different regions but also enhance our understanding of how specific objects can define the character and utility of public spaces globally. This comparative analysis of object distributions contributes significantly to constructing region-specific sidewalks' simulation environments.

\clearpage

\noindent
\textbf{Open-vocabulary search.}
To effectively retrieve digital assets corresponding to the object description pool, we developed a robust pipeline utilizing the Objaverse~\citep{objaverse} and ObjaverseXL~\citep{objaverseXL} repositories, known for their extensive digital assets. The process begins with the extraction of digital assets using a multi-threaded approach for further processing. Each downloaded asset is then rendered into 20 distinct images, capturing various angles to provide a comprehensive visual representation. Following~\citep{luo2023scalable,luo2024view}, viewpoints with higher quality are used for the calculation of visual feature embedding.

For the matching process, we leverage the BLIP2 \citep{li2023blip2} model, a pre-trained feature extractor, to align visual data with our textual descriptions. This involves processing the images to extract visual features and concurrently transforming textual descriptions into embeddings. These embeddings are compared using cosine similarity to determine the semantic correspondence between text and images, allowing us to identify and collect the digital assets that best match the descriptions.

Once the assets are collected, a meticulous review process is initiated for each category. We manually inspect each asset, filtering out those that are of low resolution, lack realism or do not meet our quality standards. The selected assets are then uploaded into MetaUrban to adjust asset characteristics such as size, position, and orientation. This meticulous curation ensures that only high-quality digital assets are incorporated into our static object dataset.

\noindent
\textbf{Object repository extension.}
MetaDrive provides an interface for including objects enabled by recent advances in 3D content generation, such as 3D object reconstruction~\citep{liu2023zero,kerbl20233d} and generation~\citep{poole2023dreamfusion,chen2023text}. Thus, one can easily further extend the object repository with generated contents. Also, this function can work together with scene customization (Section~\ref{scene_customization}) to get customized scenes with specific objects.

\subsection{Cohabitant Populating}

\noindent
\textbf{Appearances.}
We include 1,100 3D human models, 5 kinds of vulnerable road users -- bikers, skateboarders, scooter riders, and electric wheelchair users, and 6 kinds of mobile machines as cohabitants in the MetaUrban simulator. The number of dynamic agents in a scene can be set by the parameters respectively. The environment initialization time and RAM usage are only proportional to the number of individual agents. For example, 100 same agents will take the same initialization time and RAM usage as one. This schema can be used to significantly increase the maximum number of spawned agents for a specific hardware.

\noindent
\textbf{Movements.}
We include 3 daily movements -- idle, walking, and running, as well as 2,311 unique movements from the BEDLAM~\citep{black2023bedlam} dataset. All of the motion sequences are trimmed and checked by designers one by one to ensure their quality. With the same skeletal binding, all of the unique movements can be transferred to all of the 3D human models directly. Thus, we can get 1,100 $\times$ 2,311 numbers of human-motion pairs.

\noindent
\textbf{Trajectories.}
\label{trajectory}
We harness ORCA~\citep{van2011reciprocal} and Push and Rotate (P\&R) algorithm~\citep{de2014push} to get the trajectories of all dynamic agents. First, we build the 0-1 mask that indicates whether the grid is a walkable region or not. Then, we sample the start and ending points for each agent randomly, followed by generating their 2D trajectories by using the model of ORCA~\citep{van2011reciprocal} and P\&R~\citep{de2014push}. The trajectory plan process is efficient, running within 5s for 100 agents on a Core i9 CPU processor. Vehicles will also be added in dynamic scenes. All traffic vehicles will follow IDM policies, as MetaDrive~\citep{li2022metadrive} does.

\subsection{Scene Customization}
\label{scene_customization}

MetaUrban supplies various compositional elements, such as street blocks, objects, pedestrians, vulnerable road users, and other mobile agents’ appearances and dynamics. With just a few simple lines of specification, it is easy to create customized urban spaces of interest, such as street corners, plazas, and parks.

\subsection{User Interface}

\begin{figure*}[h!]
    \centering
    \includegraphics[width=0.9\linewidth]{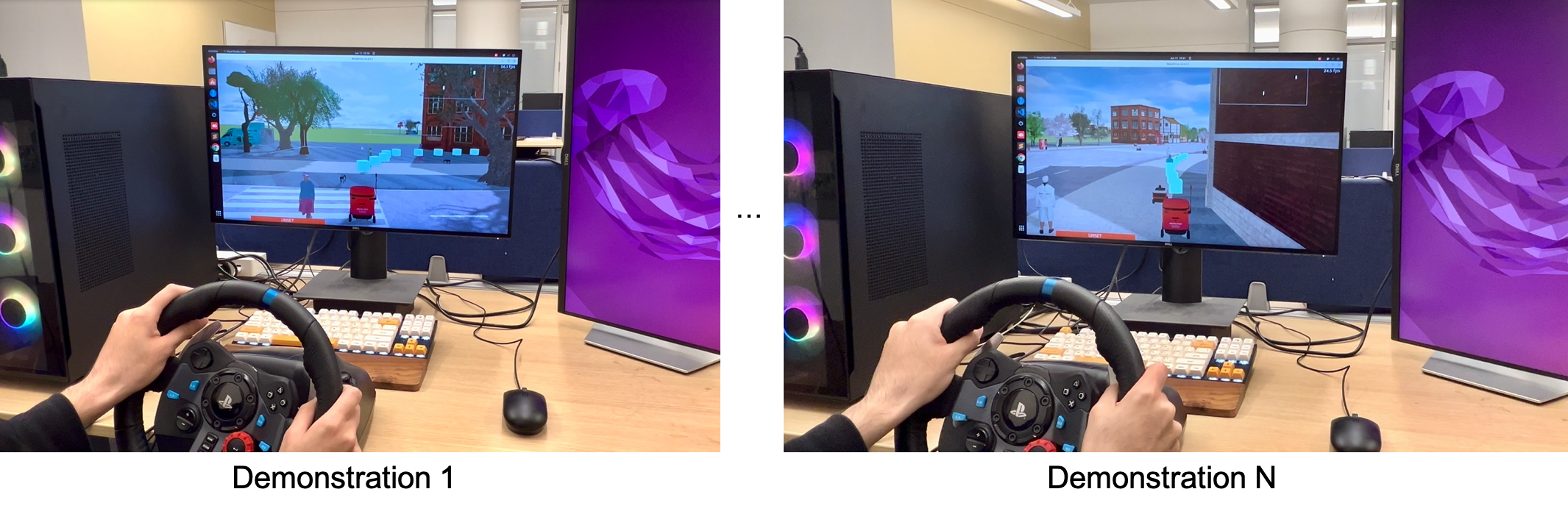}
    \caption{\textbf{Demonstration data collection with the user interface.}
    }
    \label{fig:interface}
\end{figure*}

MetaUrban provides user interfaces for two purposes: 1) Demonstration data collection for Offline RL and IL. 2) Object labeling and scene customization. 
For demonstration data collection, MetaUrban provides interfaces for mouse, keyboard, joystick, and racing wheel. We can easily collect human expert demonstrations as shown in Figure~\ref{fig:interface}.
In addition, MetaUrban provides tools for object labeling -- size, orientation, and attributes, and scene customization -- assigning the locations of the selected objects.

\subsection{Simulator Comparison}
\label{sec:simulator_comparison}

We will compare MetaUrban with other simulators below in Table~\ref{tab:comparisons}, through the scale, sensor, and feature dimensions.
For the scale, MetaUrban can generate infinite scenes with a procedural generation pipeline. It provides the largest number of humans (1,100) and movements (2,314) among all simulation environments.
For objects, so far, we have provided 10,000. Compared to other simulators, all of the objects from MetaUrban are urban-specific. Also, we provide an interface to extend object data to any size easily with recent advances in 3D content generation (Section~\ref{retrieval}). 
For the sensor, MetaUrban provides RGBD, semantic, and lidar.
For the feature, different from other simulators, MetaUrban provides real-world distribution of the object's categories and uses a more sophisticated path plan algorithm to get the natural agent's trajectories. It also provides flexible user interfaces -- mouse, keyboard, joystick, and racing wheel, which vastly ease the collection of human expert demonstration data.
MetaUrban uses PyBullet as its physical engine, which is open-source and highly accurate in physics simulation, providing a cost-effective and flexible solution for future developments.
MetaUrban uses Panda3D~\citep{goslin2004panda3d} for rendering, which is a lightweight, open-source game engine with seamless Python integration, providing a flexible and accessible development environment.

\begin{table}[h!]
    \caption{\textbf{Comparison of Embodied AI simulators.} We compare MetaUrban to simulators specialized for three environments -- indoor, driving, and social navigation environments.}
    \label{tab:comparisons}
    \centering
    \resizebox{1.0\textwidth}{!}{
    \begin{tabular}{l cccc c cccc c ccccc}
        \toprule
        & \multicolumn{4}{c}{\textbf{Scale}} && \multicolumn{4}{c}{\textbf{Sensor}} && \multicolumn{5}{c}{\textbf{Feature}} \\
        \cmidrule{2-5}
        \cmidrule{7-10}
        \cmidrule{12-16}
        Simulator & \makecell{\# of\\Scenes} & \makecell{\# of\\Objects} & \makecell{\# of\\Rigged Humans} & \makecell{\# of\\Human Motions} && RGBD & Semantic & LiDAR & Acoustic && \makecell{Object Category\\Distribution} & \makecell{Env. Agent\\Trajectory} & User Interface & Physics Engine & Scenario \\
        \midrule
        HuNavSim~\citep{perez2023hunavsim} & 5& \xmark & 5& 6&& \xmark & \xmark & \xmark & \xmark && \xmark & Social Force& \xmark &  Gazebo& Social \\
        SEAN 2.0~\citep{tsoi2022sean} & 3& 34 & <100 & 1 && \cmark & \xmark & \xmark & \xmark && Manual & Social Force& \xmark &  Unity& Social \\
        SocNavBench~\citep{biswas2022socnavbench} & 4& \xmark & \xmark & \xmark && \cmark & \xmark & \xmark & \xmark && \xmark & \xmark & \xmark &  \xmark & Social \\
        \midrule
        SUMO~\citep{krajzewicz2002sumo} & $\infty$ & \xmark & \xmark & \xmark && \xmark & \xmark & \xmark & \xmark && \xmark & \xmark& \xmark&  \xmark& Driving \\
        CARLA~\citep{dosovitskiy2017carla} & 15& 66,599 & 49& 1&& \cmark & \cmark & \cmark & \cmark && Manual& Rule-based& Keyboard, Joystick&  Unreal4& Driving \\
        MetaDrive~\citep{li2022metadrive} & $\infty$ & 5 & 1 & 1 && \cmark & \cmark & \cmark & \cmark && Manual & Rule-based & \cmark &  PyBullet& Driving \\
        \midrule
        AI2-THOR~\citep{kolve2017ai2} & 120 & 609 & \xmark & \xmark && \cmark & \cmark & \xmark & \cmark && Manual & \xmark & Mouse & Unity & Indoor \\
        ThreeDWorld~\citep{gan2020threedworld} & 15 & 200 & \xmark & \xmark && \cmark & \cmark & \xmark & \cmark && Manual & \xmark & VR & Flex & Indoor \\
        iGibson 2.0~\citep{li2021igibson} & 15 & 1,217 & \xmark & \xmark && \cmark & \cmark & \xmark & \xmark && Manual & \xmark & Mouse, VR & PyBullet & Indoor \\
        ProcTHOR~\citep{deitke2022procthor} & $\infty$ & 1,547 & \xmark & \xmark && \cmark & \cmark & \xmark & \cmark && Manual & \xmark & \xmark & Unity & Indoor \\
        OmniGibson~\citep{li2024behavior} & 306 & 5,215 & \xmark & \xmark && \cmark & \cmark & \cmark & \xmark && Manual & \xmark & \xmark & PhysX & Indoor \\
        Habitat 3.0~\citep{puig2023habitat} & 211 & 18,656 & 12 & 3 && \cmark & \cmark & \xmark & \xmark && Manual & Rule-based & Mouse, Keyboard, VR & Bullet & Indoor \\
        \midrule
        \multirow{2}{*}{\textbf{MetaUrban}} & \multirow{2}{*}{\textbf{$\infty$}} & \multirow{2}{*}{\textbf{10,000}} & \multirow{2}{*}{\textbf{1,100}} & \multirow{2}{*}{\textbf{2,314}} && \multirow{2}{*}{\textbf{\cmark}} & \multirow{2}{*}{\textbf{\cmark}} & \multirow{2}{*}{\textbf{\cmark}} & \multirow{2}{*}{\textbf{\xmark}} && \multirow{2}{*}{\textbf{Real-world}} & \textbf{ORCA} & \textbf{Mouse, Keyboard} & \multirow{2}{*}{\textbf{PyBullet}} & \multirow{2}{*}{\textbf{Urban}} \\
    &&&&&&&&&&&&\textbf{+P$\&$R}&\textbf{Joystick, Racing Wheel}\\
        \bottomrule
    \end{tabular}
    }
\end{table}

\section{MetaUrban-12K Dataset}
\label{sec:metaurban_dataset}

\paragraph{Data.} 
Based on the MetaUrban simulator, we construct the MetaUrban-12K dataset, including 12,800 interactive urban scenes for training (MetaUrban-train) and 1,000 scenes for testing (MetaUrban-test).
For the train and test sets, we sample randomly from the 6 templates (a-f) of sidewalks shown in Figure~\ref{fig:ground_plan} (right) with the same distributions of objects and dynamics.
We further construct an unseen test set (MetaUrban-unseen) with 100 scenes for zero-shot experiments, in which we sample from the unseen template (g) -- Wide Commercial Sidewalk, unseen objects, trajectories of agents with further designers' manual adjustments according to real-world scenes. In addition, to enable the fine-tuning experiments, we construct a training set of 1,000 scenes with the same distribution of MetaUrban-unseen, termed MetaUrban-finetune.
12K scenes can be generated in 12 hours on a local workstation. Notably, our MetaUrban platform can easily extend the scale of urban scenes from a multi-block level to a whole city level.
To enable the Offline RL and IL training, we collect expert demonstration data from a well-trained RL agent and human operators, forming 30,000 steps of high-quality demonstration data. The success rate of the demonstration data is 60$\%$, which can be taken as a reference for the experiments of Offline RL and IL.

\begin{figure*}[h!]
    \centering
    \includegraphics[width=1\linewidth]{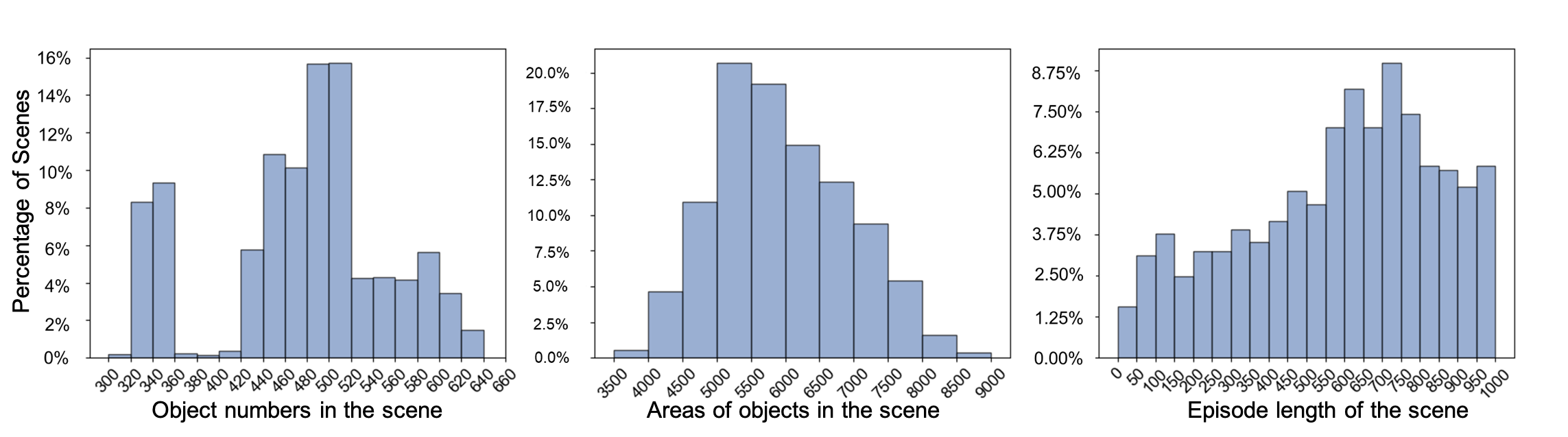}
    \caption{\textbf{MetaUrban-12K statistics.} (Left) Distribution of object numbers in the scene. (Middle) Distribution of areas occupied by objects in the scene. (Right) Distribution of episode length in the scene.
    }
    \label{fig:Statistics}
\end{figure*}

\paragraph{Statistics.}
Figure~\ref{fig:Statistics} shows distributions of the number of objects (left), areas occupied by objects (middle), and episode length (right).
As shown in the distribution of object numbers, there are lots of objects in each scenario with a minimal value of 300.
As shown in the distribution of objects' areas, objects in the dataset cover large areas, which complies with a normal distribution centered at 5250$m^2$. 
As shown in the distribution of episode length, more than 20$\%$ of them are more than 800 steps.
From these distributions, we can observe that scenes are significantly challenging in MetaUrban-12K for agents to navigate through, which are crowded and with long horizons.

\section{Experiment Details}
\label{sec:exp_detail}

This section discusses the settings of environments, action spaces, observation spaces, evaluation metrics, training details for methods, as well as the reward and cost in the benchmarks of Point Navigation (PointNav) and Social Navigation (SocialNav), respectively.

\subsection{PointNav Experiments}

\paragraph{Environments.} For PointNav experiments, there are only static objects besides the ego agent in the environment. To evaluate the trained policy, we split seven types of sidewalks into six types for training and validation with one for test. The one used for the test is the Wide Commercial Sidewalk, in which the frontage zone buffer will be, as well as some unseen objects.

We use delivery bots as the ego agent in our experiments. The task of agents in PointNav experiments is following the trajectory in the environment that navigates from start points to ending points, ensuring that it does not collide with other objects.
To generate such a task, we harness ORCA~\citep{van2011reciprocal} and Push and Rotate (P\&R) algorithm~\citep{de2014push} to get the trajectory of the ego agent after placing objects. The process is the same as discussed in Section~\ref{trajectory}. Notably, there may be some trajectories with small moving distances, we set a threshold of 5$m$ to filter out scenarios with small moving distances for testing to evaluate different methods more effectively.

\paragraph{Action spaces.} We use the same continuous action space as MetaDrive~\citep{li2022metadrive}, which is a 2-dimensional vector that normalized to $[-1.0, 1.0]$ indicating the acceleration and steering rate of the agent. Considering that the dynamics of a delivery bot is different from a vehicle, we change some core parameters like maximum velocity, maximum acceleration, maximum steering rate.

\paragraph{Observation spaces.} Multi-modal observations are provided by MetaUrban, including RGB, Depth, Semantic Map, and LidAR. We use LidAR in all of our experiments for its 3D information of the surrounding environment, which provides distance and direction of the nearest object within a $50m$ maximum detecting distance centering at the ego.
  
\paragraph{Evaluation metrics.} For PointNav, an episode is considered successful if the agent issues the DONE action, defined as completing 95\% of the set route within 1,000 maximum steps. The agent is evaluated using the Success Rate (SR) and Success weighted by Path Length (SPL)~\citep{anderson2018evaluation,batra2020objectnav} metrics, which measure the effectiveness and efficiency of the path taken by the agent.
Additionally, to measure the safety performance of the trained policy, we define the cost function by two events, \textit{i.e.}, crashing with objects on the sidewalk or buildings in the building zone.  $+1$ cost is given once those events occur.

 \paragraph{Methods.}
In our study, we employ a diverse set of 7 baseline models to establish comprehensive benchmarks on MetaUrban. These models span various domains, including Reinforcement Learning, Safe Reinforcement Learning, Offline Reinforcement Learning, and Imitation Learning.

\noindent
\textit{Reinforcement learning.}
In the realm of Reinforcement Learning, we use the Proximal Policy Optimization (PPO) \citep{schulman2017proximal} for evaluation. PPO is a widely adopted and effective method that strikes a balance between sample complexity and ease of tuning, and it is easy to scale as it adopts parallel and distributed training well. The agent in this setting is trained to maximize the reward, which we carefully design to encapsulate the desired behavior of the agent in the MetaUrban environment. The specifics of the reward structure will be discussed in the subsequent paragraph. We train the PPO using the same set of hyperparameters with 128 parallel environments, which occupy 128 processes. The total training time is 12 hours, and 5M environment steps for PointNav on a single Nvidia A5000 GPU. The detailed hyperparameters are provided in Table~\ref{tab:ppo-hyper}.

\begin{table}[!h]
    \centering
    \caption{Hyper-parameters of RL and SafeRL for PointNav.}
    \vspace{0.1in}
    \begin{tabular}{l l} \toprule  
    \textbf{PPO/PPO-Lag/PPO-ET Hyper-parameters} & \textbf{Value} \\
    \hline 
    Environmental horizon $T$ &1,000\\
         Learning rate&5e-5\\
 Discount factor $\gamma$&0.99\\
 GAE parameter $\lambda$&0.95\\
  Clip parameter $\epsilon$ & 0.2\\
         Train batch size & 25,600 \\
 SGD minibatch size&256\\
         Value loss coefficient&1.0\\ 
         Entropy loss coefficient&0.0\\  \midrule
         Cost limit &1 \\ \bottomrule
    \end{tabular}
    \label{tab:ppo-hyper}
\end{table}

\noindent
\textit{Safe reinforcement learning.} As driving in urban spaces is a safety-critical application, it is important to evaluate Safe Reinforcement Learning (SafeRL) algorithms. In the domain of SafeRL, we utilize two approaches: PPO with a Lagrangian constraint (PPO-Lag)~\citep{ray2019benchmarking} and PPO with modeling of Early Terminated Markov Decision Processes (PPO-ET)~\citep{sun2021safe}. Both methods aim to ensure that the learned policies adhere to specific safety constraints while optimizing the reward. PPO-Lag incorporates a Lagrangian term into the objective function to enforce the constraints, while PPO-ET changes the modeling of the Constrained Markov Decision Process (CMDP) to a new unconstrained MDP, the optimal policy that coincidences with the original CMDP.

For PPO-Lag~\citep{ray2019benchmarking}, it considers the learning objectivate as Equation~\ref{ppo-lag} rather than adding negative cost as rewards.
\begin{equation}
    \label{ppo-lag}
    \max_{\theta}\min_{\lambda\geq0} E_{\tau}[R_{\theta}(\tau)-\lambda(C_{\theta}(\tau)-d)]
\end{equation}
where $R_{\theta}$, $C_{\theta}$, $\theta$, and $d$ are episodic reward, episodic cost, parameters of the policy, and given cost threshold, respectively.

The rule for PPO-ET~\citep{sun2021safe} is to stop when the constraint cost exceeds a given value, which can be easily implemented in practice. 

We implement both of these SafeRL methods based on OmniSafe~\citep{ji2023omnisafe}. We train both of them with 50 parallel environments and the training takes 12 hours for PointNav on a single Nvidia A5000 GPU. The detailed hyperparameters are provided in Table~\ref{tab:ppo-hyper}. 

\noindent
\textit{Offline reinforcement learning.}
For Offline Reinforcement Learning, we employ two prominent methods: Implicit Q-Learning (IQL) \citep{kostrikov2021offline} and Twin Delayed Deep Deterministic Policy Gradient with Behavior Cloning (TD3+BC)~\citep{fujimoto2021minimalist}. We create the dataset for PointNav by combining $20\%$ human demonstrations with $80\%$ demonstrations from a well-trained PPO policy, consisting of 30,000 samples with approximately 60\% success rate. The training is purely offline and takes around 2 hours on a single Nvidia A5000 GPU for 100 epochs. The detailed hyperparameters for IQL and TD3+BC are provided in Table~\ref{tab:iql-hyper} and \ref{tab:tdbc-hyper}, respectively.

\begin{table}[!h]
    \centering
    \caption{Hyper-parameters of IQL.}
    \vspace{0.1in}
    \begin{tabular}{l l} \toprule 
    \textbf{IQL Hyper-parameters} & \textbf{Value} \\
    \midrule 
         Learning rate&  1e-4\\
         Discount factor $\gamma$&  0.99\\
         Target critic update ratio&  5e-3\\
 Inverse temperature $\beta$&3.0\\
 Log std range&(-5.0, 2.0)\\
 Expectile&0.7\\ \bottomrule
    \end{tabular}
    \label{tab:iql-hyper}
\end{table}

\begin{table}[!h]
    \centering
    \caption{Hyper-parameters of TD3+BC.}
    \vspace{0.1in}
    \begin{tabular}{l l} \toprule 
    \textbf{TD3+BC Hyper-parameters} & \textbf{Value} \\
    \midrule 
         Learning rate&  1e-4\\
         Discount factor $\gamma$&  0.99\\
         Target critic update ratio& 5e-3\\
         Actor update delay& 2\\
         BC loss coefficient& 2.5\\ \bottomrule
    \end{tabular}
    \label{tab:tdbc-hyper}
\end{table}

\noindent
\textit{Imitation learning.}
For Imitation Learning algorithms, we use the same high-quality mixed demonstration used in Offline Reinforcement Learning. In the Imitation Learning setting, the agent learns to mimic the behavior shown in the expert demonstration, and it is differentiated from Offline Reinforcement Learning in the sense that the agent does not have access to the rewards. We employ two well-established methods: Behavior Cloning (BC) \citep{michael1995bc} and Generative Adversarial Imitation Learning (GAIL) \citep{ho2016generative}. BC is a straightforward approach that trains the agent to directly match the actions of the expert given the observed states. GAIL, on the other hand, formulates the imitation learning problem as a two-player game between the agent and a discriminator, which tries to distinguish between the agent's behavior and the expert's demonstrations. The detailed hyperparameters for IQL and TD3+BC are provided in table~\ref{tab:bc-hyper} and~\ref{tab:gail-hyper}, respectively.

\begin{table}[!h]
    \centering
    \caption{Hyper-parameters of BC.}
    \vspace{0.1in}
    \begin{tabular}{l l} \toprule 
    \textbf{BC Hyper-parameters} & \textbf{Value} \\
    \midrule 
    Dataset size & 30,000 \\
         Learning rate&  1e-4\\
         SGD batch size&64\\
         SGD epoch &40\\
          
          \bottomrule
    \end{tabular}
    \label{tab:bc-hyper}
\end{table}

\begin{table}[!h]
    \centering
    \caption{Hyper-parameters of GAIL.}
    \vspace{0.1in}
    \begin{tabular}{l l} \toprule 
    \textbf{GAIL Hyper-parameters} & \textbf{Value} \\
    \midrule 
    Dataset size & 30,000 \\
    SGD batch size&64\\
    Sample batch size&12,800\\
    Generator Learning rate&  1e-4\\
    Discriminator Learning rate&  3e-3\\  
    Generator optimization epoch&  5\\
    Discriminator optimization epoch&  2,000\\   
    Clip parameter $\epsilon$ & 0.2 \\ 
          \bottomrule
    \end{tabular}
    \label{tab:gail-hyper}
\end{table}

\paragraph{Reward and cost.}
The reward function is composed as follows:
\begin{equation}
    R = R_{term} + c_1 R_{disp} + c_2 R_{lateral} + c_3 R_{steering} + c_4 R_{crash}
\end{equation}
Specifically, 
\begin{itemize}
\item Terminal reward $R_{term}$: 
 a sparse reward set to $+5$ if the vehicle reaches the destination, and $-5$ for out of route. If given $R_{term}\neq 0$ at any time step $t$, the episode will be terminated at $t$ immediately. 

 \item Displacement reward $R_{disp}$: a dense reward defined as $R_{disp}=d_t-d_{t-1}$, wherein the $d_t$ and $d_{1}$ denote the longitudinal position of the ego agent in Frenet coordinates of current lane at time $t$ and $t-1$, respectively. We set the weight of $R_{disp}$ as $c_1=0.5$.

  \item Lateral reward $R_{lateral}$: a dense reward defined as $R_{lateral}=-||l_t||$, wherein the $l_t$ denotes the lateral offset of the ego agent in Frenet coordinates of current lane at time $t$, which is designed to prevent agent driving on  non walkable areas. We set the weight of $R_{lateral}$ as $c_2=1.0$.

   \item Steering smoothness reward $R_{steering}$: a dense reward defined as $R_{steering}=-||s_t-s_{t-1}||\cdot v_t$, wherein the $s_t$ and $s_{t-1}$ denotes the steering of the agent at $t$ and $t-1$, respectively. And $v_t$ denotes the speed of the agent at time $t$. This reward term is designed as a regularization to prevent the agent changing the steering too frequently. We set the weight of $R_{steering}$ as $c_3=0.1$.

   \item Crash reward $R_{crash}$: a dense negative reward defined as $-1(c_{t})$, wherein the $c_{t}$ denotes the collision between agents and any other objects at time $t$ and $1(\cdot)$ is the indicator function. It's notable we do not use the termination strategy for collision as in MetaDrive~\citep{li2022metadrive}. We set the weight of $R_{crash}$ as $c_4=1.0$.
    
\end{itemize}

And for benchmarking Safe RL algorithms, collision to any objects raises a cost $+1$ at each time step.

\subsection{SocialNav Experiments}

\paragraph{Environments.}
For SocialNav experiments, most settings are the same as the ones in PointNav. The most important difference is that dynamic agents will also be present in the environment. The trajectories of environmental agents are generated together by using the model of ORCA~\citep{van2011reciprocal} with P\&R~\citep{de2014push}. Since vehicles are inherited from MetaDrive~\citep{li2022metadrive}, we use the same parameter to control its density, \textit{i.e.}, traffic density $0.05$ in our experiments.

\paragraph{Evaluation metrics.} For SocialNav, an episode is considered successful if the agent issues the DONE action, defined as completing 95\% of the set route within 1,000 maximum steps. The agent is evaluated using the Success Rate (SR) and Social Navigation Score (SNS)~\citep{deitke2022retrospectives}, which is the average of Success weighted by Time Length (STL) and Personal Space Compliance (PSC). SNS measures the agent in terms of safety and efficiency.

\paragraph{Methods.} We benchmark the same methods as in PointNav experiments with the same hyperparameters. However, due to the involvement of lots of dynamic agents, the training speed of SocialNav is about approximately 1/3 of PointNav on online methods. The cost scheme is defined as raising a cost of $+1$ at each time step if the ego agent crashes with any agents, vehicles, or objects.

\subsection{Evaluation across Mobile Machines}
\label{sec:policy_execution}

In this experiment, we evaluate the influence of mechanical structures in \textit{policy execution} on different terrains. We conduct experiments on three types of wheeled mobile machines -- a delivery bot, an electric wheelchair, and a mobility scooter, with remarkably different specifications, such as wheelbase, wheel radius, and wheel width.
We designed three sufficiently long runways with three kinds of terrains: slopes, stairs, and roughs. From the starting point, each runway has a gradually increasing difficulty. Slopes will have an increasingly steeper angle; stairs will have increasingly higher step heights; roughs will have increasingly larger bumps. We apply the same "moving forward" policy to each mobile machine to test the longest distance and time duration they can travel before termination, and report the metrics of $x$-displacement (m) and Time to fall (s)~\citep{agarwal2023legged} respectively. Terminal conditions are getting stuck, slowing down significantly, and toppling over.

Results are shown in Table~\ref{tab:polity_execution}.
The mobility scooter, which has the largest wheelbase, wheel width, and radius, achieves the best performance in the slopes test. It indicates that a larger wheelbase increases stability and reduces the risk of tipping backward on steep inclines, while wheel width and radius help in better traction on slopes.
All three machines show similar but poor performance in the stairs test. It indicates the inherent defect of wheeled mobile machines and emphasizes the importance of accessibility in public urban spaces.
The delivery bot, which has the smallest wheelbase, wheel width, and radius, achieves poor performance on all three tests. It indicates that although the delivery bot's structure gives it good maneuverability on flat surfaces, it comes at the cost of losing stability on complex terrains.

\begin{table}[h]
\small
\centering
\caption{\textbf{Evaluation of policy execution across mobile machines.} For each row of different terrains, \legendsquare{colorbest} indicates the best performance among the three machines.} 
\label{tab:polity_execution}
\begin{center}
\resizebox{1\columnwidth}{!}{
    \begin{tabular}{c|ccc|ccc}
        \toprule
        \multirow{2}{*}{\textbf{Terrain}}& \multicolumn{3}{c|}{\textbf{$x$-displacement (m) $\uparrow$}} & \multicolumn{3}{c}{\textbf{Time to fall (s) $\uparrow$}}\\
        \cmidrule{2-7}
        & \makecell{\textbf{Delivery Bot}} & 
        \makecell{\textbf{Wheelchair}} &
        \makecell{\textbf{Mobility Scooter}} &
        \makecell{\textbf{Delivery Bot}} & 
        \makecell{\textbf{Wheelchair}} &
        \makecell{\textbf{Mobility Scooter}}\\
        \toprule
        Wheelbase ($m$) &0.45 &0.5 &0.6 &0.45 &0.5 &0.6 \\
        Wheel radius ($m$) &0.1 &0.15 &0.2 &0.1 &0.15 &0.2 \\
        Wheel width ($m$) &0.1 &0.1 &0.15 &0.1 &0.1 &0.15 \\
        \midrule
        Slopes &31.90 &34.58 &\best{38.07} &6.3 &6.9 &\best{7.7} \\
        Stairs &38.55 &\best{38.94} &38.67 &7.0 &\best{7.8} &7.2 \\
        Roughs &28.06 &31.91 &\best{34.17} &5.8 &6.4 &\best{6.7} \\
        \bottomrule
    \end{tabular}
}
\end{center}
\end{table}

\section{Unique Challenges in Urban Micromobility}
\label{sec:unique_challenges}

In this section, we delve into and validate four unique challenges a mobile machine will encounter in public urban space, which is the stage of urban micromobility tasks, distinct from previous indoor and driving environments, \textit{i.e.}, long horizon tasks in large-scale scenes, multifarious terrains, diverse obstacles, and dense pedestrians.

\noindent
\textbf{Long horizon tasks in large-scale scenes.}
In urban spaces, mobile machines need to perform long-horizon tasks with large-scale scenes connected with several street blocks, which are several orders of magnitude larger than indoor environments. We compare the performance of two models trained with different episode length settings. Setting-1: mean length with 10$m$ (a common length following the indoor environment ProcThor~\citep{deitke2022procthor}). Setting-2: mean length with 410$m$ (a common length in urban micromobility tasks). Both models are then tested on the MetaUrban-test dataset.
As shown in Table~\ref{tab:unique_challenge} (Left), the model trained with Setting-1 achieves poor performances when testing on the urban environments. However, when trained in an urban environment with a longer episode length (Setting-2), the model's performance improves dramatically.
It indicates that long-horizon tasks in large-scale scenes bring a unique challenge to mobile machines, and validate the necessity of procedural generation of large-scale urban scenes in MetaUrban.

\begin{table}[h]
\small
\caption{\textbf{Unique challenges validation.} (Left) Long horizon tasks in large-scale scenes. (Right) Dense pedestrians.}
\label{tab:unique_challenge}
  \centering
  \begin{minipage}{0.3\textwidth} 
    \centering
    \begin{tabular}{c|cc}
        \hline
        &\textbf{Setting-1}  & \textbf{Setting-2}  \\
        \hline
        SR$\uparrow$ & 10\%  &41\%  \\
        SPL$\uparrow$ & 0.08  &0.38  \\
        \hline
    \end{tabular}
    \label{tab:length}
  \end{minipage}
  \hspace{0.05\textwidth}
  \begin{minipage}{0.6\textwidth}
    \centering
    \begin{tabular}{c|cccc}
        \hline
        & \textbf{Setting-1} & \textbf{Setting-2} & \textbf{Setting-3} & \textbf{Setting-4} \\
        \hline
        SR$\uparrow$ &24\% &16\% &10\% &8\% \\
        Cost$\downarrow$ &0.51 &0.75 &0.74 &0.68 \\
        \hline
    \end{tabular}
    \label{tab:pede}
  \end{minipage}
\end{table}

\noindent
\textbf{Multifarious terrains.}
In indoor environments, most of the grounds are smooth and even, while in driving environments, roadway surfaces can only have slight cracks and damage.
Yet, in public urban spaces, mobile machines will encounter multifarious complex terrains, such as slopes, stairs, and roughs. As shown in Table~\ref{tab:polity_execution}, different terrains will deeply influence the performance of mobile machines with different mechanical structures, such as wheelbase, wheel radius, and width, \textit{etc.}. Results show that along with the increasing difficulty of terrains, all of the machines will fail because of getting stuck, barely moving, or toppling over.
It indicates that multifarious terrains bring a unique challenge to mobile machines, and validate the necessity of the terrain generation system designed in MetaUrban.

\begin{figure*}[h!]
    \centering
    \vspace{-0.12in}
    \includegraphics[width=1\linewidth]{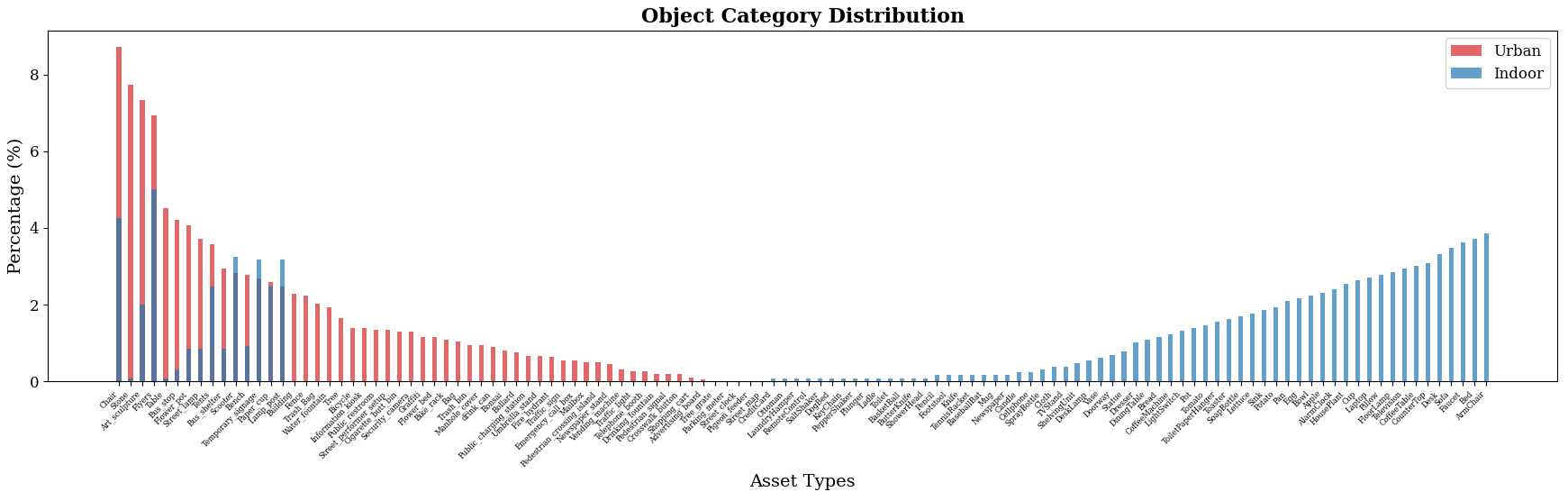}
    \caption{\textbf{Object category distributions.} Red: Distribution of object category in urban spaces. Blue: Distribution of object category in indoor spaces.
    }
    \label{fig:object_distribution}
\end{figure*}

\noindent
\textbf{Diverse obstacles.}
In indoor environments, even though there are many objects, the distribution has a large variation compared to urban spaces. In driving environments, there are only a few obstacles on roadways, such as cones and barriers.
As shown in Figure~\ref{fig:object_distribution}, we compare the distribution of object category in MetaUrban with that in ProcThor~\citep{deitke2022procthor}, a state-of-the-art indoor simulation environment. We can observe a significant variation in object category distributions between public urban spaces and indoor spaces, although these two scenarios both accommodate a lot of objects.
The diversity, particularity, and concentration of obstacles in urban spaces present a unique challenge for mobile machines. The statistical results validate the necessity of the pipeline of scalable obstacle filling in MetaUrban.

\noindent
\textbf{Dense pedestrians.}
In indoor environments, humans will share walkable spaces with humans; however, there are only 3-5 people in one room in common (as shown in Habitat 3.0~\citep{puig2023habitat}). In driving environments, except the intersections, there exist barely any shared spaces for pedestrians and vehicles. In contrast, in public urban spaces, almost all of the spaces for mobile machines are shared with pedestrians.
We compare models trained with different pedestrian densities and locations. Setting-1: 10 pedestrians per 100-meter episode (a common scenario in an urban environment). Setting-2: 5 pedestrians per 100-meter episode (a common scenario in an indoor environment). Setting-3: 10 pedestrians per 100-meter episode but only shown in intersections (a common scenario in a driving environment). All the three models are then tested on the MetaUrban-test dataset. Setting 4: using the model trained in Setting-1 and testing on a density of 30 pedestrians per 100-meter episode (a crowd scenario in an urban environment).

Results are shown in Table~\ref{tab:unique_challenge} (Right). We take Setting-1 as a reference. With fewer pedestrians in Setting-2, the success rate will decrease, and the cost will increase significantly, indicating a higher frequency of bumping into pedestrians. With different distribution but the same pedestrian density in Setting-3, both success rate and cost degrade dramatically, indicating the unique challenge of sharing walkable regions with pedestrians. In Setting 4, we further increase the pedestrian's density and see a huge degradation in success rate but a moderate degradation in Cost. It indicates that having more pedestrians poses a significant challenge for the agent to reach the goal point. Interestingly, the agent still attempts to avoid pedestrians due to its effective training in public urban spaces.
Results demonstrate that the high pedestrian density and interaction frequency in public urban spaces will place a unique challenge for mobile machines. It also validates the importance of the Cohabitant Populating module in MetaUrban.

\clearpage

\section{Datasheet}
\label{sec:data_sheet}

\begin{longtable}{p{0.35\linewidth} |p{0.6\linewidth} }
    \bottomrule
        \multicolumn{2}{c}{\rule{0in}{0.2in}\textbf{Motivation}\vspace{0.10in}}\\
    \toprule
     For what purpose was the dataset created? & The dataset was created to enable agents training on diverse scenes and facilitate AI-driven urban micromobility research.\\[0.15in]
     \midrule
     Who created and funded the dataset? &
      This work was created and funded by the MetaUrban team at the University of California, Los Angeles.\\
     
    \bottomrule
        \multicolumn{2}{c}{\rule{0in}{0.2in}\textbf{Composition}\vspace{0.10in}}\\
    \toprule
    What do the instances that comprise the dataset represent? & Each instance is a JSON file including the configuration of our MetaUrban environment and a specific seed.

    \\[0.15in]
    \midrule
    How many instances are there in total (of each type, if appropriate)? & There are 12,800 urban scenes released in the MetaUrban-12K dataset, along with the code to sample substantially more. \\[0.15in]
    \midrule
    Does the dataset contain all possible instances or is it a sample (not necessarily random) of instances from a larger set? & We offer 12,800 urban scenes, with the ability to generate more using procedural generation scripts.\\[0.15in]
    \midrule
    What data does each instance consist of? & Each scene is specified as a JSON file including the configuration of our MetaUrban environment and a specific seed.\\[0.15in]
    \midrule
    Is there a label or target associated with each instance? & No.\\[0.15in]
    \midrule
    Is any information missing from individual instances? & No.\\[0.15in]
    \midrule
    Are relationships between individual instances made explicit (e.g., users' movie ratings, social network links)? & Each urban scene is created independently, so there are no connections between the scenes.\\[0.15in]
    \midrule
    Are there recommended data splits? & Yes. See Section 4 in the main paper.\\[0.15in]
    \midrule
    Are there any errors, sources of noise, or redundancies in the dataset? & No.\\[0.15in]
    \midrule
    Is the dataset self-contained, or does it link to or otherwise rely on external resources (e.g., websites, tweets, other datasets)? & The dataset is self-contained.\\[0.15in]
    \midrule
    Does the dataset contain data that might be considered confidential? & No.\\[0.15in]
    \midrule
    Does the dataset contain data that, if viewed directly, might be offensive, insulting, threatening, or might otherwise cause anxiety? & No.\\
    \bottomrule
        \multicolumn{2}{c}{\rule{0in}{0.2in}\textbf{Collection Process}\vspace{0.10in}}\\
    \toprule
    How was the data associated with each instance acquired? & Each scene was procedurally generated.\\[0.15in]
    \midrule
    If the dataset is a sample from a larger set, what was the sampling strategy? & The dataset consists of 12,800 scenes, each by sampling the parameters of its composed elements.\\[0.15in]
    \midrule
    Who was involved in the data collection process? & The authors were the sole individuals responsible for creating the dataset.\\[0.15in]
    \midrule
    Over what timeframe was the data collected? & Data was collected in Sept. 2024.\\[0.15in]
    \midrule
    Were any ethical review processes conducted? & No.\\
    \bottomrule
        \multicolumn{2}{c}{\rule{0in}{0.2in}\textbf{Preprocessing/Cleaning/Labeling}\vspace{0.10in}}\\
    \toprule
    Was any preprocessing/cleaning/labeling of the data done? & We label each object's location area and pivots to make them spawn in target functional zones and face a natural direction.\newline

    We use VLMs to automatically label 2D images of cities worldwide, which enables the extraction of real-world category distribution of objects in urban spaces.\\[0.15in]
    \midrule
    Was the ``raw'' data saved in addition to the preprocessed/cleaned/labeled data? & There is no raw data.\\[0.15in]
    \midrule
    Is the software that was used to preprocess/clean/label the data available? & The code related to preprocessing, cleaning, and labeling the data will be made available.\\
    \bottomrule
        \multicolumn{2}{c}{\rule{0in}{0.2in}\textbf{Uses}\vspace{0.10in}}\\
    \toprule
    Has the dataset been used for any tasks already? & Yes. See Section 4 of the main paper.\\[0.15in]
    \midrule
    What (other) tasks could the dataset be used for? & The scenes can be used in a wide variety of tasks in urban micromobility, embodied AI, computer vision, and urban planning. \newline
    \\[0.15in]
    \midrule
    Is there anything about the composition of the dataset or the way it was collected and preprocessed/cleaned/labeled that might impact future uses? & No.\\[0.15in]
    \midrule
    Are there tasks for which the dataset should not be used? & Our dataset can be used for both commercial and non-commercial purposes.\\
    \bottomrule
        \multicolumn{2}{c}{\rule{0in}{0.2in}\textbf{Distribution}\vspace{0.10in}}\\
    \toprule
    Will the dataset be distributed to third parties outside of the entity on behalf of which the dataset was created? & Yes. We plan to make the entirety of the work open-source, including the code used to generate scenes and train agents, the scripts to get the MetaUrban-12K dataset, and the asset repositories.\\[0.15in]
    \midrule
    How will the dataset be distributed? & The scene files will be distributed with a custom Python package.\newline
    
    The code, asset, and repositories will be distributed on GitHub.\\[0.15in]
    \midrule
    Will the dataset be distributed under a copyright or other intellectual property (IP) license, and/or under applicable terms of use (ToU)? & The scene dataset, 3D asset repository, and code will be released under the Apache 2.0 license. \\[0.15in]
    \midrule
    Have any third parties imposed IP-based or other restrictions on the data associated with the instances? & For 3D human assets, we use Synbody~\citep{yang2023synbody}. Its license is CC BY-NC-SA 4.0. For movement sequences, we use BEDLAM~\citep{black2023bedlam}. See \url{https://bedlam.is.tue.mpg.de/license.html} for its license. \\[0.15in]
    \midrule
    Do any export controls or other regulatory restrictions apply to the dataset or to individual instances? & No.\\
    \bottomrule
        \multicolumn{2}{c}{\rule{0in}{0.2in}\textbf{Maintenance}\vspace{0.10in}}\\
    \toprule
    Who will be supporting/hosting/maintaining the dataset? & The authors will be providing support, hosting, and maintaining the dataset.\\[0.15in]
    \midrule
    How can the owner/curator/manager of the dataset be contacted? & For inquiries, email <metaurban\_team@gmail.com>.\\[0.15in]
    \midrule
    Is there an erratum? & We will use GitHub issues to track issues with the dataset.\\[0.15in]
    \midrule
    Will the dataset be updated? & We will continue adding support for new features to make the urban scenes more diverse and realistic. We also intend to support new tasks in the future.\\[0.15in]
    \midrule
    If the dataset relates to people, are there applicable limits on the retention of the data associated with the instances (e.g., were the individuals in question told that their data would be retained for a fixed period of time and then deleted)? & The dataset does not relate to people. \\[0.15in]
    \midrule
    Will older versions of the dataset continue to be supported/hosted/maintained? & Yes. Revision history will be available for older versions of the dataset.\\[0.15in]
    \midrule
    If others want to extend/augment/build on/contribute to the dataset, is there a mechanism for them to do so? & Yes. The work will be open-sourced, and we intend to offer support to assist others in using and building upon the dataset.\\[0.15in]
\bottomrule
    \caption{A datasheet \citep{Gebru2021DatasheetsFD} for MetaUrban and MetaUrban-12K.}
\end{longtable}

\newpage

\section{Performance}
\label{sec:performance}

We measure the performance of MetaUrban under varying street blocks, different densities of static objects, and dynamic agents in the scene. All experiments are conducted on a single Nvidia V100 GPU and in a single process. For the environment, there are approximately 200 objects covering $1500m^2$ on average. We sample 1,000 steps for actions and run 10 times to report the average and standard error results of FPS. For the RGB and depth image, we use the 128$\times$128 resolution. On average, for the RGB, Depth, and LiDAR observation, we achieve 50$\pm$15, 60$\pm$10, and 120$\pm$12 FPS in training, respectively.

\section{Robustness}
\label{sec:robustness}

We trained PPO on PointNav with different seeds and found that the variance of the performance across different seeds is small. The success rate of the PPO agents is $0.695 \pm 0.014$ on the MetaUrban-test set and $0.638 \pm 0.060$ on the MetaUrban-unseen set.

\section{Discussion}
\label{sec:discussion}

\noindent
\textbf{Impact.}
As the first urban space simulator, MetaUrban could benefit broad areas across Embodied AI, Economy, and Society.
1) \textit{Embodied AI.} MetaUrban contributes to advancing areas such as robot navigation, social robotics, and interactive systems. It could facilitate the development of robust AI systems capable of understanding and navigating complex urban environments.
2) \textit{Economy.} MetaUrban could be used in businesses and services operating in urban environments, such as last-mile food delivery, assistive wheelchairs, and trash-cleaning robots. It could also drive innovation in urban planning and infrastructure development by providing simulation tools and insights into how spaces are utilized, thereby enhancing the economic and societal efficiency of public urban spaces like sidewalks and parks.
3) \textit{Society.} By enabling the safe integration of robots and AI systems in public spaces, MetaUrban could support the development of assistive technologies that can aid in accessibility and public services. Using AI in public spaces might foster new forms of social interaction and community services, making urban spaces more livable and joyful.
4) \textit{Potential negative societal impacts.} The integration of AI and robots in urban environments, while beneficial, raises several concerns. Increased surveillance could infringe on privacy, while automation may lead to job displacement and exacerbate economic inequalities. Societal dependency on technology poses risks of dysfunction during failures, and the presence of robots might alter social norms and interactions. Thus, the environmental impact of manufacturing and operating urban simulators must be carefully managed. Addressing these issues is crucial for ensuring that the benefits of such technologies are realized without detrimental societal consequences.

\noindent
\textbf{Limitations.}
1) \textit{Real-world scene distribution.} In this work, we extract object category distribution from real-world data of urban spaces. Other than the real-world distribution of object categories, the distribution of object location and scene layout is also important for constructing specialized scenes for agent training. Extraction of such distribution relies on an accurate reconstruction of 3D scenes from real-world videos or even images, and thus is extremely challenging.
An interesting direction is extracting real-world scene distribution from in-the-wild videos, including object category, object location, and scene layout. Then, we can build a digital twin of a target scene for the agent's training. It could help to develop scene-specific agents.
2) \textit{Interactive agent behaviors.} In this work, we construct the environmental agents‘ dynamic with deterministic methods, determining their movements and trajectories with rules. However, in the real world, all environmental agents are interactive; their behaviors are affected by each other and the surrounding environments. An interesting research direction is to endow personal traits like job, personality, and purpose to agents and harness the advances of LLMs~\citep{achiam2023gpt} and LVMs~\citep{liu2023visual} to form social~\citep{puig2023nopa} and interactive behaviors~\citep{park2023generative} of agents in urban scenes spontaneously.
3) \textit{Robots' additional capability learning.} In urban micromobility, safe navigation through the city is the primary goal for mobile machines.
However, additional capabilities, such as locomotion and manipulation, can enable robots to perform more complex tasks in urban spaces. Thus, an important direction is to extend MetaUrban to support additional capabilities learning gradually. It could enable various complex but important services in urban environments.
4) \textit{Efficiency.} In this work, different from indoor scenes and driving simulators, MetaUrban supports generating complex interactive urban scenes with arbitrary scales. However, with the increase in scale, the number of objects and dynamic agents will surge dramatically, which will bring the degradation of the efficiency of physical simulation and rendering. A promising direction is to integrate more sophisticated physical engines and renders.

\noindent
\textbf{Sim-to-Real.}
MetaUrban is an embodied AI simulator that can enable fast iterative model development and evaluation before deployment.
However, as it is impossible to perfectly replicate the real world without some loss, sim-to-real gap is an \textit{inevitable issue} faced by all embodied AI simulators~\citep{kolve2017ai2,savva2019habitat,shen2021igibson,li2024behavior,szot2021habitat,deitke2020robothor,deitke2022procthor,krajzewicz2002sumo,li2022metadrive,dosovitskiy2017carla}, which also exists in MetaUrban.
Following the standard setting embodied AI simulators, \textit{i.e.}, AI2-THOR~\citep{kolve2017ai2}, Habitat~\citep{savva2019habitat}, and ProcTHOR~\citep{deitke2022procthor}, we did not focus on real-world experiments so far.

However, positioning MetaUrban as a long-term maintained and sustainable community infrastructure, we do think the sim-to-real gap is a fundamental issue for simulators and are continuously working in two directions to tackle this problem.
1) \textit{Building real-world evaluation platform.} Exceeding from an embodied AI platform, we are constructing an end-to-end experimental platform that spans from the MetaUrban simulation to the real-world deployment of robots, to progressively improve the designs in the simulator. We have conducted experiments in Unitree’s Go2 quadruped robot and COCO Robotics’s wheeled robot. We found that using more abstract observations (such as semantic maps and LiDAR maps provided in MetaUrban) to train robots with domain randomization can already achieve good generalizability to real-world environments.
2) \textit{Improving the realism of the simulation environment.} First, to improve appearance realism, we are working on integrating Nvidia OmniVerse as the renderer. Then, to improve physical realism, we are working on integrating Nvidia PhysX into MetaUrban.

\noindent
\textbf{Future work.}
1) \textit{Foundation model.} MetaUrban can easily generate infinite urban scenes with a large quantity of semantics and complex interactions, which could facilitate the pre-training of foundation models (like LLMs and LVMs) that can be used for downstream agent learning tasks. 
2) \textit{Human-robot cohabitate.} Mobile machines have started emerging in the urban space, which makes it no longer exclusive to humans. We plan to work with urban sociologists to study the influence of robots on human urban life through both simulation and field experiments.
3) \textit{Improve limitations.} The directions discussed in limitations and sim-to-real gaps are also meaningful future work we will conduct.
In summary, MetaUrban, as a new urban environment simulator, will bring a lot of new interesting research directions. We are dedicated to maintaining MetaUrban in the long term and supporting the community's efforts to develop it into a sustainable infrastructure.

\end{document}